%% file: main.tex
\documentclass[10pt,twocolumn,letterpaper]{article}

\usepackage{cvpr}              

\usepackage{graphicx}
\usepackage{amsmath}
\usepackage{amssymb}
\usepackage[inline]{enumitem}
\usepackage{booktabs}
\usepackage{pifont}
\usepackage[flushleft]{threeparttable}

\usepackage{xspace}
\newcommand{\Method}{SCALE\xspace}
\newcommand{\FullName}{\underline{S}elf-Supervised \underline{C}ontr\underline{A}stive Lifelong \underline{LE}arning without Prior Knowledge}
\newcommand{\C}{\checkmark}
\newcommand{\X}{$\times$}

\DeclareMathOperator*{\argmin}{arg\,min}

\usepackage[pagebackref,breaklinks,colorlinks]{hyperref}

\usepackage[capitalize]{cleveref}
\crefname{section}{Sec.}{Secs.}
\Crefname{section}{Section}{Sections}
\Crefname{table}{Table}{Tables}
\crefname{table}{Tab.}{Tabs.}

\def\cvprPaperID{37} 
\def\confName{CLVision}
\def\confYear{2023}

\begin{document}

\title{SCALE: Online Self-Supervised Lifelong Learning without Prior Knowledge}

\author{
Xiaofan Yu\textsuperscript{\rm 1}, 
Yunhui Guo\textsuperscript{2},
Sicun Gao\textsuperscript{1},
Tajana Rosing\textsuperscript{1} \\
\textsuperscript{\rm 1} University of California San Diego,
\textsuperscript{\rm 2} University of Texas at Dallas \\
{\tt\small \{x1yu, sicung, tajana\}@ucsd.edu} \\ 
{\tt\small yunhui.guo@utdallas.edu}
}

\maketitle

\begin{abstract}
Unsupervised lifelong learning refers to the ability to learn over time while memorizing previous patterns without supervision. Although great progress has been made in this direction, existing work often assumes strong prior knowledge about the incoming data (e.g., knowing the class boundaries), which can be impossible to obtain in complex and unpredictable environments. In this paper, motivated by real-world scenarios, we propose a more practical problem setting called online self-supervised lifelong learning without prior knowledge.
The proposed setting is challenging due to the non-\textit{iid} and single-pass data, the absence of external supervision, and no prior knowledge. 
To address the challenges, we propose \FullName~(\Method) which can extract and memorize representations on the fly purely from the data continuum.
\Method is designed around three major components: a pseudo-supervised contrastive loss, a self-supervised forgetting loss, and an online memory update for uniform subset selection. All three components are designed to work collaboratively to maximize learning performance.
We perform comprehensive experiments of \Method under \textit{iid} and four \textit{non-iid} data streams. The results show that \Method outperforms the state-of-the-art algorithm in all settings with improvements up to 3.83\%, 2.77\% and 5.86\% in terms of $k$NN accuracy on CIFAR-10, CIFAR-100, and TinyImageNet datasets.
We release the implementation at \url{https://github.com/Orienfish/SCALE}.
\end{abstract}

\vspace{-6mm}
\section{Introduction}
\label{sec:intro}

Lifelong learning, or continual learning, refers to the ability to continuously learn over time by acquiring new knowledge and consolidating past experiences. One major challenge of lifelong learning is to combat \textit{catastrophic forgetting}, i.e., updating the model using new samples degrades existing knowledge learned in the past~\cite{mccloskey1989catastrophic,goodfellow2013empirical}. 

Existing work has assumed various levels of \emph{prior knowledge} about the input data stream. 
\textit{Supervised Lifelong Learning} presumes the presence of task and class labels along with samples~\cite{kirkpatrick2017overcoming,lopezpaz2017gradient,chaudhry2018efficient,guo2020improved}. \textit{General Continual Learning} or \textit{task-free continual learning} eliminates the task labels and boundaries to focus on real-time adaptation to non-stationary continuum with limited memory, but still using class labels~\cite{aljundi2019task,buzzega2020dark,ijcai2022p0446,ye2022task}. \textit{Unsupervised Lifelong Learning} completely removes all labels; therefore, the algorithm needs to distill the knowledge from raw samples or streaming structure on its own~\cite{jiang2017variational,achille2018life,wu2018memory}.

\begin{figure}[t]
\begin{center}
\centerline{\includegraphics[width=0.46\textwidth]{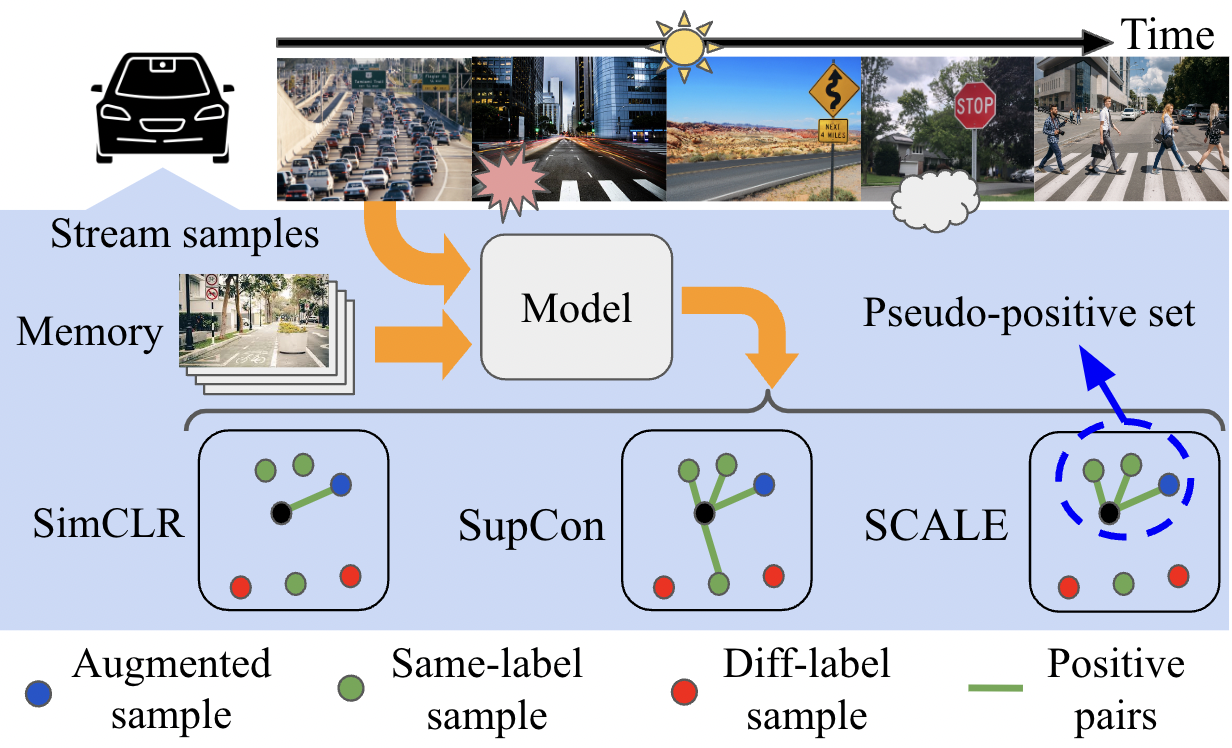}}
\vspace{-3mm}
\caption{\small \Method functions on a self-driving vehicle where the order of the input image sequence can be unforeseeable due to environmental or operational factors. \Method learns self-supervisedly by contrasting with memory samples. \Method's pseudo-contrastive loss is inspired from the InfoNCE objective~\cite{oord2018representation}. While SimCLR~\cite{chen2020simple} only uses an augmented sample and SupCon~\cite{khosla2020supervised} uses samples with the same label to form a positive set, to improve similarity within the set, \Method self-distills a pseudo-positive set based on pairwise similarity. \Method does not rely on any supervision or prior knowledge.}
\label{fig:motivation}
\end{center}
\vspace{-12mm}
\end{figure}

While great progress has been made in lifelong learning, it is still challenging to deploy the existing algorithms in the wild to learn over time. One of the reasons is that even in the pure unsupervised setting, existing works assumed knowing the class boundary or the total number of classes in advance~\cite{rao2019continual,ijcai2022p0483,pratama2022unsupervised}.
Such prior knowledge greatly eases the difficulty of learning without forgetting. For example, if the class boundary is distinct and known, the learning algorithm can expand the network or create a new memory buffer whenever detecting a class shift. But these prior knowledge is extremely difficult, if not impossible, to obtain in real-world environments which are \textit{complex} and \textit{unpredictable}.
Specifically, consider a camera mounted on a vehicle and an application of continuously training an image classification algorithm as the vehicle moves around (Figure~\ref{fig:motivation}). 
The sequence of incoming samples depends on the environment and the trajectory of the vehicle, hence, is very hard to predict when and how smooth the shift is.

In this paper, to align with the unpredictable real-world scenarios, we extend the current unsupervised learning setting to a more challenging and practical case: online unsupervised lifelong learning without prior knowledge. 
In particular, we make no assumption on the input streams:
\vspace{-2mm}
\begin{enumerate}[label=(\roman*),itemsep=-1.5mm]
    \item Unlike offline self-supervised learning~\cite{chen2020simple,caron2020unsupervised}, the input data is non-\textit{iid} and \textit{single-pass}, i.e., all data samples appear only once.
    \item Unlike \textit{General Continual Learning}~\cite{buzzega2020dark,ijcai2022p0446} and task-based lifelong learning~\cite{he2021unsupervised,fini2021self,lin2021continual,madaan2022representational}, the class and task labels are not given (\textit{no external supervision}).
    \item Unlike VAE-based design~\cite{rao2019continual} and KMeans-based progressive clustering~\cite{he2021unsupervised,taufique2022unsupervised}, the task or class boundaries and the number of classes are unknown in advance (\textit{no prior knowledge}).
\end{enumerate}
\vspace{-2mm}
Additionally, the input stream can have distinct/blurred class boundaries or an imbalanced class appearance, all of which are not revealed to the algorithm. 
Our problem setting reflects the complexity and difficulty of lifelong learning problems in the real world\footnote{In this paper we focus on image classification while the same setup and methodology can be easily extended to other applications as well.}.

Recognizing the unique challenges, we propose \FullName~(\Method).
\Method is designed around three major components: 
a pseudo-supervised contrastive loss for contrastive learning, a self-supervised forgetting loss for lifelong learning, and an online memory update for uniform subset selection. All components are critical to the final learning performance: the contrastive loss enhances the similarity relationship by contrasting with memory samples, the forgetting loss prevents catastrophic forgetting, and the memory buffer retains the most ``representative'' raw samples within the limited buffer size. Our loss functions utilize pairwise similarity among the feature representations, thus eliminating the dependency on labels or prior knowledge. Moreover, contrastively learned representations have been shown to be more robust against catastrophic forgetting compared to the use of end-to-end cross-entropy loss~\cite{cha2021co2l}.

Our \textbf{contributions} can be summarized as follows:
\begin{enumerate}[label=(\arabic*),itemsep=-1.3mm]
\item We propose a more practical setting for unsupervised lifelong learning which assumes that the input data streams are non-\textit{iid} and single pass, and no external supervision or prior knowledge is given.
\item We design \Method to extract and memorize knowledge on-the-fly without supervision and prior knowledge. \Method uses contrastive lifelong learning based on self-distilled pairwise similarity, along with an online memory update to retain the ``representative'' raw samples on imbalanced streams.
\item We perform comprehensive experiments on five different types of single-pass data stream sampled from CIFAR-10, CIFAR-100 and TinyImageNet datasets. \Method outperforms state-of-the-art algorithms in all settings.
\end{enumerate}

\vspace{-3mm}
\section{Related Work}
\label{sec:related-work}

\bgroup
\def\arraystretch{1.2}
\begin{table*}[tb]
\small
\centering
\caption{\small Comparison of previous work and \Method (this paper) on assumed prior knowledge.}
\vspace{-2mm}
\label{tbl:problem}
\begin{threeparttable}[t]
\centering
\begin{tabular}{c|cccc} 
\toprule
\small
Papers & Single-pass & Non-\textit{iid} & No task labels & No class labels  \\ \hline 
VASE~\cite{achille2018life}, CURL~\cite{rao2019continual}, L-VAEGAN~\cite{ye2020learning} & \X & \C & \C & \C \\ 
He \textit{et al.}~\cite{he2021unsupervised}, CCSL~\cite{lin2021continual}, CaSSLe~\cite{fini2021self}, LUMP~\cite{madaan2022representational}
 & \C & \C & \X & \C \\ 
Tiezzi \textit{et al.}~\cite{ijcai2022p0483}, KIERA~\cite{pratama2022unsupervised} & \C & \C & \C & \X \\ 
STAM~\cite{smith2019unsupervised}, \Method (this paper) & \C & \C & \C & \C \\ 
 \bottomrule
\end{tabular}
\end{threeparttable}
\vspace{-4mm}
\end{table*}

\textbf{Self-Supervised Learning (SSL)} has been developed to learn low-dimensional representations on offline datasets without class labels, for various downstream tasks.
Variational autoencoder (VAE)-based designs aimed for data reconstruction assuming various prior models in the latent space~\cite{nalisnick2016stick,jiang2017variational,joo2020dirichlet}.
Progressive clustering-based methods alternated between network update and clustering for self-labeling until convergence~\cite{xie2016unsupervised,caron2018deep,haeusser2018associative,chang2017deep,rebuffi2020lsd,caron2020unsupervised}. 
Information theory-based techniques maximized the mutual information between representations of augmented samples to retain invariance and avoid degenerate solutions~\cite{ji2019invariant,hu2017learning,zbontar2021barlow,ermolov2021whitening,bardes2021vicreg,li2021self}.
Contrastive learning draws closer the augmented representation pairs while pushing away the others~\cite{oord2018representation,chen2020simple,he2020momentum,chen2020improved,chen2020big}.
Recent architecture techniques such as BYOL, SimSiam and OBoW~\cite{grill2020bootstrap,chen2021exploring,gidaris2020online} used asymmetric networks to prevent learning trivial representations.
However, all the above-mentioned works are designed for offline \textit{iid} data and do not address catastrophic forgetting.

\textbf{Supervised Lifelong Learning} has been widely explored in three lines: dynamic architecture~\cite{rusu2016progressive,ostapenko2019learning,von2020continual,rajasegaran2020itaml,abati2020conditional,Lee2020A}, regularization~\cite{kirkpatrick2017overcoming,zenke2017continual,aljundi2018memory,ritter2018online,ahn2019uncertainty,yu2020semantic,zhang2020class}, and experience replay using a memory buffer~\cite{rebuffi2017icarl,lopezpaz2017gradient,chaudhry2018efficient,buzzega2020dark,guo2020improved,chrysakis2020online,hu2021continual,wang2022memory,tiwari2022gcr}.
Recently, a large amount of effort has been invested in online supervised lifelong learning. Most works used memory replay, such as
Co2L~\cite{cha2021co2l}, 
CoPE~\cite{de2021continual}, 
GMED~\cite{jin2021gradient}, 
DualNet~\cite{pham2021dualnet}, 
ASER~\cite{shim2021online}, 
SCR~\cite{mai2021supervised}, 
OCM~\cite{guo2022online},
ODDL~\cite{ye2022task},
OCD-Net~\cite{ijcai2022p0446}.
Nevertheless, the problem is significantly simplified with the presence of class labels.

\textbf{Unsupervised Lifelong learning (ULL)} is mostly studied under offline \textit{iid} data with multiple passes on the entire dataset during training~\cite{jiang2017variational,achille2018life,wu2018memory,ye2020learning}.
In contrast, online ULL is more challenging due to the non-\textit{iid} and single-pass data continuum.
Lifelong generative models leveraged mixture generative replay to mitigate catastrophic forgetting during online updates~\cite{rao2019continual,ramapuram2020lifelong}. However, these VAE-based methods were computationally expensive.
Many recent works have applied self-supervised knowledge distillation on task-based online ULL.
\textbf{He \textit{et al.}}~\cite{he2021unsupervised} utilized pseudo-labels from KMeans clustering to guide knowledge preservation from the previous task. 
\textbf{CCSL}~\cite{lin2021continual} employed self-supervised contrastive learning for intra- and inter-task distillation.
\textbf{CaSSLe}~\cite{fini2021self} proposed a general framework for SSL backbones, which extracted the best possible representations that are invariant to task shifts.
\textbf{LUMP}~\cite{madaan2022representational} mitigated forgetting by interpolating the current task's samples with the finite memory buffer.
But all of these works relied on task boundaries to generate good results.
\textbf{Tiezzi \textit{et al.}}~\cite{ijcai2022p0483} developed a human-like attention mechanism for continuous video streams with little supervision.
\textbf{KIERA}~\cite{pratama2022unsupervised} and \textbf{STAM}~\cite{smith2019unsupervised} employed expandable memory architecture for single-pass data using online clustering, novelty detection and memory update. KIERA required labeled samples in the initial batch of each task for cluster association. 
The problem definition of STAM is most similar to ours. Yet, STAM's memory architecture cannot be trained with common optimizers, and thus is limited in fine-tuning for downstream tasks. 

We summarize the existing contributions for online ULL in Table~\ref{tbl:problem} based on the assumed prior knowledge.
The proposed \Method excels existing works in that \Method learns low-dimension representations online without any external supervision or prior knowledge about task, class or data; thus, it better adapts unpredictable real-world environments.

\begin{figure*}[ht]
\begin{center}
\vspace{-2mm}
\centerline{\includegraphics[width=0.93\textwidth]{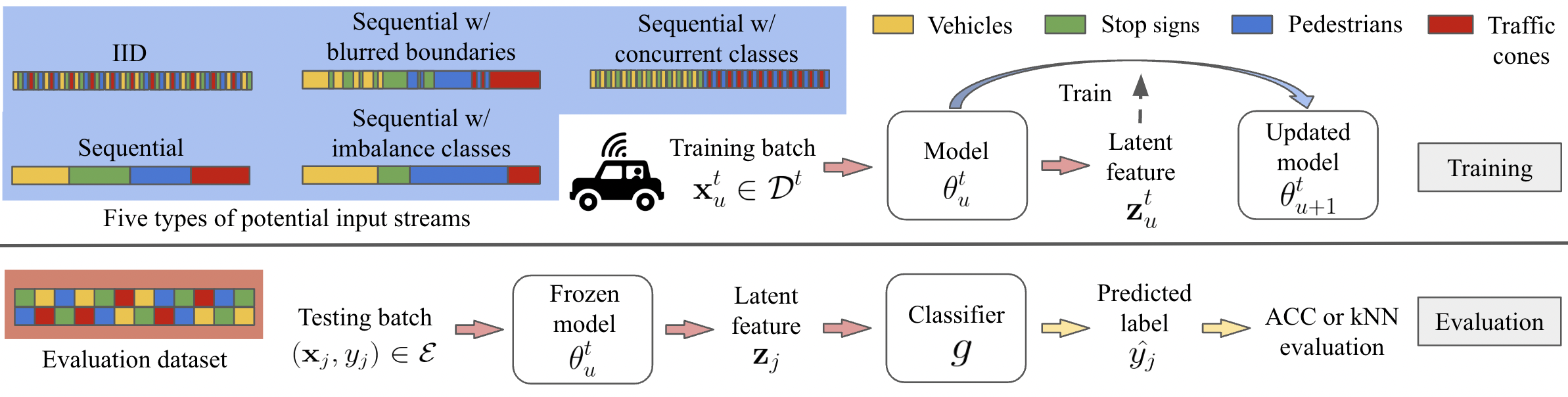}}
\vspace{-4mm}
\caption{\small The training and evaluation setup of online ULL taking the self-driving vehicle as an example. The input samples are from classes of vehicles, stop signs, pedestrians and traffic cones. We consider five typical input streams of \textit{iid}, sequential classes, sequential classes with blurred boundaries, imbalanced sequential classes, and sequential classes with concurrent class appearance. 
Each training step updates the model self-supervisedly based on the feature representations, while periodic evaluation is triggered on a separate evaluation dataset using supervised or unsupervised classifiers on the learned feature representations.}
\label{fig:scenario}
\end{center}
\vspace{-11mm}
\end{figure*}

\vspace{-2mm}
\section{Online Unsupervised Lifelong Learning without Prior Knowledge}
\label{sec:problem}




In this section, we present the online unsupervised lifelong learning problem without prior knowledge. Our setup is motivated by real-world applications and extended from previous studies by removing certain assumptions.


\noindent \textbf{Input streams.} 
We assume the data comes in a class- (or distribution-) incremental manner. 
Such a setup mimics continuous and periodic sampling while the surrounding environment changes over time.
Suppose that the input samples are drawn from a sequence of $T$ classes with each class corresponding to a unique distribution in $\left \{ \mathcal{P}^1, ..., \mathcal{P}^T\right \}$. 
The complete input sequence can then be represented as $\mathcal{D} = \left \{ \mathcal{D}^1, ...  ,\mathcal{D}^T\right \}$ where $\mathcal{D}^t$ denotes a series of $n_t$ batches of samples, i.e., $\mathcal{D}^t = \{ X_1^t,...,X_{n_t}^t \}$. With $t$ denoting the class ID and $u$ representing the batch ID in the current class, each batch of data $X^t_u$ is a set of samples $\{ \mathbf{x}_1^t,...,\mathbf{x}_{|X^t_u|}^t \}$, where $\mathbf{x}_i^t \sim \mathcal{P}^t(\mathbf{X})$.
In the rest of the paper, we use capital letters to denote batches and lowercase letters for individual samples.
Each training batch $X_u^t \in \mathcal{D}^t$ appears once in the entire stream (single-pass) while the task and class labels are not revealed.
The total number of classes $T$, the transition boundaries and the batch numbers $n_t$ are not known by the learning algorithm either.
Our goal is to learn a model that distinguishes classes or distributions $\left \{ \mathcal{P}^1, ..., \mathcal{P}^T\right \}$ \textit{at any moment} throughout the stream, without supervision by external labels or prior knowledge. 

Based on previous problem formulations~\cite{rao2019continual,smith2019unsupervised,pratama2022unsupervised}, five particular types of input streams are considered:
\begin{enumerate*}[label=(\roman*)]
    \item \textit{iid} data that is sampled \textit{iid} from all classes. 
    \item Sequential class-incremental stream where the observed classes are balanced in length and are introduced one-by-one with clear boundaries that are not known by the algorithm.
    \item Sequential class-incremental stream with blurred boundaries. The boundary is blurred by mixing the samples from two consecutive classes, mimicking class shift is smooth and difficult to detect.
    \item Imbalanced sequential class-incremental stream uses different batch sizes in each class, mimicking distribution shifts at unpredictable times.
    \item Sequential class-incremental stream with concurrent classes where more than one class is incrementally introduced at the time. In this paper two classes are revealed concurrently and $\mathcal{P}^i$ refers to their combined distribution.
\end{enumerate*}
To aid understanding, we use a self-driving vehicle with a mounted camera as an example to visualize all five input streams as shown in Figure~\ref{fig:scenario}.


\noindent \textbf{Training and evaluation protocol.}
The training and evaluation setup is similar to~\cite{rao2019continual,smith2019unsupervised} and is detailed in Figure~\ref{fig:scenario}.
The model is a representation mapping function to a low-dimensional feature space, i.e., $f_\theta: \mathcal{X} \rightarrow \mathcal{Z}$ where $\theta$ represents learnable parameters and $\mathcal{Z}$ refers to the low-dimensional feature space. The training proceeds self-supervisedly based on the feature representation batch $Z_u^t = f_{\theta_u^t}(X_u^t)$. 
As for evaluation, we periodically test the frozen model $\theta_u^t$ on a separate dataset $\mathcal{E} = \{ (\mathbf{x}_j, y_j) \}$ as the training progresses. 
We randomly sample an equal amount of labeled samples from each class possibly seen in $\left \{ \mathcal{P}^1, ..., \mathcal{P}^T\right \}$ and add them to $\mathcal{E}$.  Thus even when the class has not shown up in the sequence, it is always included in $\mathcal{E}$.
For each testing sample $(\mathbf{x}_j, y_j) \in \mathcal{E}$, we first compute the learned latent representations $\mathbf{z}_j = f_{\theta_u^t}(\mathbf{x}_j)$. We then apply a classifier $g: \mathcal{Z} \rightarrow \mathcal{Y}$ on $\mathbf{z}_j$ to generate the predicted labels $\hat{y_j}$. 
The classifier $g$ can be unsupervised or supervised to evaluate different aspects of the representation learning ability.
Following previous protocols~\cite{rao2019continual,smith2019unsupervised}, we use spectral clustering, an unsupervised clustering method, and employ \textit{unsupervised clustering accuracy} (ACC) as the accuracy metric.
ACC is defined as the best accuracy among all possible assignments between clusters and target labels:
\begin{equation}
    ACC = \max_\psi \frac{\sum_{j=1}^{|\mathcal{E}|}\mathbf{1} \left \{y_j = \psi(\hat{y_j})\right \}}{|\mathcal{E}|}.
\end{equation}
Here, the predicted label $\hat{y_j}$ is the cluster assignment to sample $\mathbf{x}_j$, $\psi$ ranges over all possible one-to-one mappings between $\hat{y_j}$ and $y_j$.
For supervised classification, we employ $k$-Nearest Neighbor ($k$NN) classifier. 

\noindent \textbf{Challenges.} 
The major difference between our online unsupervised lifelong learning and previous problems is the prior assumptions about the input stream. Online ULL is more challenging than previous ULL problems as shown in Table~\ref{tbl:problem} from three aspects:
\vspace{-2mm}
\begin{enumerate}[label=(C\arabic*),leftmargin=2.2\parindent,itemsep=-1.5mm]
    \item \textbf{The non-\textit{iid} and single-pass input data streams} require online knowledge distillation, which is largely different from offline self-supervised learning with \textit{iid} data and multi-pass training~\cite{chen2020simple,caron2020unsupervised,zbontar2021barlow}.
    \item \textbf{The lack of task or class labels} differs our online ULL from \textit{General Continual Learning} (with class labels)~\cite{buzzega2020dark,ijcai2022p0446} and task-based lifelong learning (with task labels)~\cite{he2021unsupervised,fini2021self,lin2021continual,madaan2022representational}. 
    The model must distill the knowledge from the stream on its own without external supervision. 
    \item \textbf{The absence of prior knowledge.} Existing ULL methods rely on class boundaries~\cite{fini2021self,lin2021continual,he2021unsupervised,madaan2022representational} or maintain and update class prototypes after detecting a shift~\cite{rao2019continual,pratama2022unsupervised}. However, these approaches do not apply when there is no prior knowledge, especially with smooth transitions, imbalanced streams or simultaneous classes as in our online ULL.
\end{enumerate}
\vspace{-2mm}

\begin{figure*}[t]
\begin{center}
\centerline{\includegraphics[width=0.78\paperwidth]{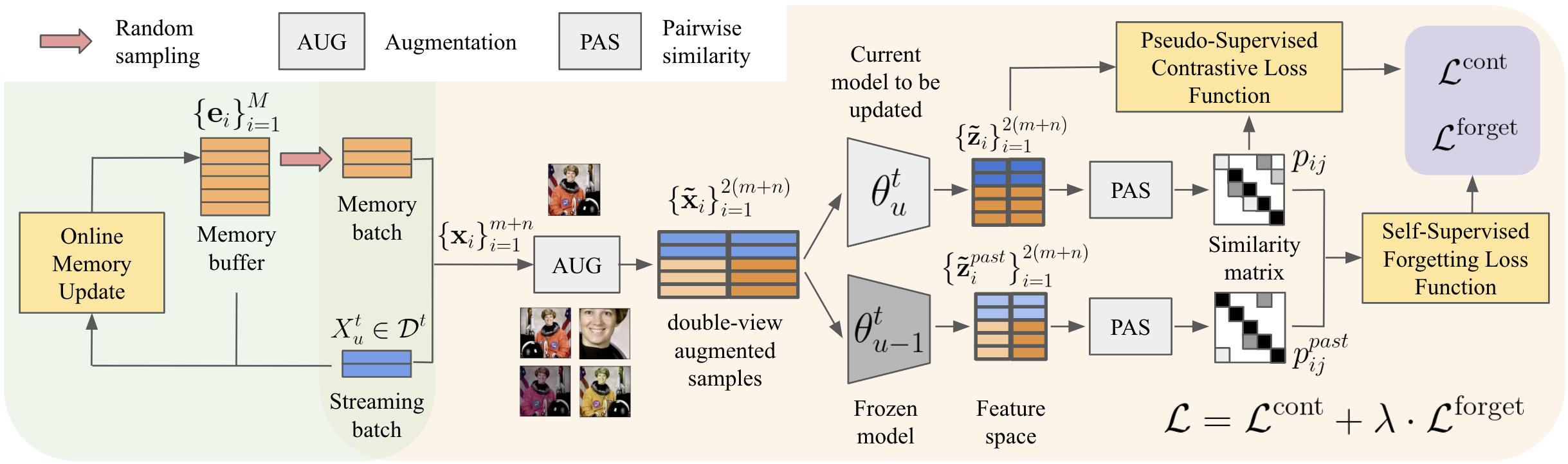}}
\vspace{-2mm}
\caption{\small The pipeline of \Method  is designed around three major components depicted by yellow boxes. The right-hand portion in orange includes the operations related to self-supervised contrastive and lifelong learning. The left-hand portion in green contains the procedures related to online memory update. \Method requires careful design for all three components to distill and memorize knowledge on the fly.}
\label{fig:method}
\end{center}
\vspace{-9mm}
\end{figure*}

\section{The Design of \Method}
\label{sec:method}




To address the above challenges of online ULL,  we propose \Method, an unsupervised lifelong learning method that can learn over time without prior knowledge. An overview of \Method is shown in Figure~\ref{fig:method}. \Method is designed around three major components (shown in yellow boxes): a pseudo-supervised contrastive loss, a self-supervised forgetting loss, and an online memory update module that emphasizes uniform subset selection.
By combining stored memory samples with the streaming samples during learning, \Method addresses challenge (C1).
Secondly, \Method uses the newly proposed pseudo-supervised contrastive learning paradigm that distills the relationship among samples via pairwise similarity. Pseudo-supervised distillation works without task or class labels thus handles challenge (C2). Learning from pairwise similarity does not depend on class boundaries or the number of classes, therefore \Method responds to challenge (C3).

We emphasize that all components are carefully designed to work collaboratively and maximize learning performance: 
the contrastive loss is responsible for extracting the similarity relationship by contrasting with memory samples, the forgetting loss retains the similarity knowledge thus prevents catastrophic forgetting, finally the online memory update maintains a memory buffer with representative raw samples in the past. 
We record the raw input samples rather than feature representations in the memory buffer because feature representations might change during training. 
The quality or the ``representativeness'' of memory samples can significantly affect learning performance, as demonstrated by our results in the evaluation section.


Figure~\ref{fig:method} shows the pipeline of \Method in detail. Memory buffer is assumed to have maximum size of $M$, and the stored memory samples are represented by $\left \{ \mathbf{e}_i\right \}_{i=1}^{M}$. 
Each streaming batch $X_u^t$ with batch size of $n = |X^t_u|$ is stacked with a randomly sampled subset of $m$ memory samples to form a combined batch $\left \{ \mathbf{x}_i\right \}_{i=1}^{m+n}$ as input to \Method.
We apply double-view augmentation to the stacked data and obtain $\left \{ \mathbf{\tilde{x}}_i\right \}_{i=1}^{2(m+n)}$ where $\mathbf{\tilde{x}}_{2k-1}, \mathbf{\tilde{x}}_{2k}$ denote two randomly augmented samples from $\mathbf{x}_k$.  
The augmented samples are fed into the representation learning model $f_\theta$ to obtain normalized low-dimensional features $\mathbf{\tilde{z}}_i = f_\theta(\mathbf{\tilde{x}}_i), \forall i \in \left \{ 1, ..., 2(m+n)\right \}$.
\Method distills pairwise similarity from $\left \{ \mathbf{\tilde{z}}_i\right \}_{i=1}^{2(m+n)}$, which are then used to compute the pseudo-supervised contrastive and forgetting losses to update the current model $\theta_u^t$.
On the other hand, online memory update takes previous memory buffer $\left \{ \mathbf{e}_i\right \}_{i=1}^{M}$ and the streaming batch $X_u^t$ as input, selects a subset of $M$ samples to store in the updated memory buffer.
We discuss the details below.



\vspace{-1mm}
\subsection{Pseudo-Supervised Contrastive Loss and Self-Supervised Forgetting Loss}
\label{sec:loss}
The loss function of \Method has two terms: a novel pseudo-supervised contrastive loss $\mathcal{L}^{\textrm{cont}}$ for learning representations and a self-supervised forgetting loss $\mathcal{L}^{\textrm{forget}}$ for preserving knowledge:
\begin{equation}
\mathcal{L} = \mathcal{L}^{\textrm{cont}} + \lambda \cdot \mathcal{L}^{\textrm{forget}}.
\end{equation}
A hyperparameter $\lambda$ is used to balance the two losses.
Both loss functions rely on pairwise similarity hence do not need prior knowledge and adapt to a variety of streams.

\noindent \textbf{Pseudo-Supervised Contrastive Loss.}
Our contrastive loss is inspired from the InfoNCE objective~\cite{oord2018representation} which enhances the similarity between positive pairs over negative pairs in the feature space. SimCLR~\cite{chen2020simple} and SupCon~\cite{khosla2020supervised} are the typical offline contrastive learning techniques using InfoNCE loss. Different from SimCLR (treats only the augmented pair as positive, unsupervised) and SupCon (forms the positive set based on labels, supervised), SCALE establishes a pseudo-positive set based on pairwise similarity.
Given a feature representation $\mathbf{\tilde{z}}_i$, its pseudo-positive pair $\mathbf{\tilde{z}}_j$ is selected from the self-distilled pseudo-positive set $\Gamma_i$. Negative pairs are all non-identical representations in the augmented batch $\left \{ \mathbf{\tilde{z}}_i\right \}_{i=1}^{2(m+n)}$. 
Formally, the pseudo-supervised contrastive loss is defined as:
\begin{equation}
\mathcal{L}^{\textrm{cont}} = \sum_{i=1}^{2n} \frac{-1}{|\Gamma_i|}\sum_{j \in \Gamma_i} \log \frac{\exp (\mathbf{\tilde{z}}_i \cdot \mathbf{\tilde{z}}_j / \tau)}{\sum_{k=1,k \neq i}^{2(m+n)} \exp (\mathbf{\tilde{z}}_i \cdot \mathbf{\tilde{z}}_k / \tau)},
\label{eq:contrastive-loss}
\end{equation}
where $\tau > 0$ is a temperature hyperparameter.
Note, that all memory samples only act as negative contrasting pairs to avoid overfitting.
Without task or class labels, \Method distills the pairwise similarity $p_{ij}$ and forms the pseudo-positive set as:
\begin{equation}
    \Gamma_i = \left \{ j \in \left \{ 1, ..., 2n\right \} | \: j \neq i, p_{ij} > \mu \right \},
\end{equation}
where $p_{ij}$ (defined later) indicates the pairwise similarity among feature representations and $\mu > 0$ is a hyperparameter as similarity threshold.
Our contrastive loss is unique and different from traditional contrastive loss functions~\cite{chen2020simple,khosla2020supervised,he2020momentum} due to the self-distilled pseudo-positive set $\Gamma_i$, which maximizes the effectiveness of unsupervised representation learning in an online setting.
\noindent \textbf{Self-Supervised Forgetting Loss.}
To combat catastrophic forgetting, we construct a self-supervised forgetting loss based on the KL divergence of the similarity distribution:
\begin{equation}
    \mathcal{L}^{\textrm{forget}} = \sum_{i=1}^{2(m+n)} \sum_{j=1, j \neq i}^{2(m+n)} - p_{ij} \cdot \log \frac{p_{ij}}{p_{ij}^{past}},
\end{equation}
where $p_{ij}, p_{ij}^{past}$ are the pairwise similarity among feature representations $\left \{ \mathbf{\tilde{z}}_i\right \}_{i=1}^{2(m+n)}$ and $\left \{ \mathbf{\tilde{z}}_i^{past}\right \}_{i=1}^{2(m+n)}$, which are mapped by the model $\theta_u^t$ and frozen model $\theta_{u-1}^t$. 
To form a valid distribution, we enforce the pairwise similarity of a given instance to sum to one: 
$\sum_{j=1, j \neq i}^{2(m+n)} p_{ij}=1, \forall i \in \left \{ 1, ..., 2(m+n)\right \}$.
The same rule applies to $p_{ij}^{past}$.
In \Method, the learned knowledge is stored by pairwise similarity. Hence penalizing the KL divergence of pairwise similarity distribution from a past model can prevent catastrophic updates.
As we are not aware of class or task boundaries, we use the frozen model from the previous batch.
Note, that a similar distillation loss is used in~\cite{cha2021co2l,he2021unsupervised,lin2021continual} but for supervised or task-based lifelong learning.

\smallskip
\noindent \textbf{Pairwise Similarity.}
Pairwise similarity is the key of \Method hence picking the suitable metric is of critical importance.
An appropriate pairwise similarity metric should (i) consider the global distribution of all streaming and memory samples, and (ii) sum to one for a given instance as required by the forgetting loss.
We adopt the symmetric SNE similarity metric from t-distributed stochastic neighbor embedding (t-SNE), which was originally proposed to visualize high-dimensional data by approximating the similarity probability distribution~\cite{van2008visualizing}:
\begin{equation}
    p_{ij} = \frac{p_{j|i} + p_{i|j}}{2}, \: p_{j|i} = \frac{\exp(\mathbf{\tilde{z}}_j \cdot \mathbf{\tilde{z}}_i / \kappa)}{\sum_{k=1, k \neq i}^{2(m+n)} \exp(\mathbf{\tilde{z}}_k \cdot \mathbf{\tilde{z}}_i / \kappa)}, \label{eq:tsne}
\end{equation}
where $\kappa > 0$ is a temperature hyperparameter.
Since the form of Equation~\eqref{eq:tsne} is similar to Equation~\eqref{eq:contrastive-loss}, in practice, the computation can be reused to improve efficiency.
The symmetric SNE similarity captures the global similarity distribution among all features without using supervision or prior knowledge.

\vspace{-1mm}
\subsection{Online Memory Update}
\label{sec:memory}
The goal of online memory update is to retain the most ``representative'' raw samples from historical streams to obtain the best outcome in contrastive learning.
One major challenge is that the input streams are non-\textit{iid} and possibly imbalanced.
Existing work has proposed various memory update strategies to extract the most informational samples, e.g., analyzing interference or gradients information~\cite{aljundi2019online,aljundi2019advances,chrysakis2020online,jin2021gradient,shim2021online}.
However, most previous works rely on class labels thus are not applicable in online ULL.
Without labels and prior knowledge, we cannot make any assumption (for example, clusters) on the manifold of the feature representations that are fed to memory update.
Purushwalkam \textit{et al.}~\cite{purushwalkam2022challenges} were the first to bring up a similar problem setting and proposed minimum redundancy (MinRed) memory update, prioritizing dissimilar samples without considering the global distribution.
Unlike MinRed, we propose to perform distribution-aware uniform subset sampling for memory update.


The input to memory update is the imbalanced combined batch $\left \{ \mathbf{x}_i \right \}_{i=1}^{M+n}$ of the previous memory buffer $\left \{ \mathbf{e}_i\right \}_{i=1}^{M}$ and streaming batch $X_u^t$.
We first map the raw samples to the feature space, i.e., $\mathbf{z}_i = f_\theta(\mathbf{x}_i), \forall i \in \left \{ 1, ..., M+n \right \}$. Then we select a subset of $M$ samples from $\left \{ \mathbf{z}_i \right \}_{i=1}^{M+n}$ and store the corresponding raw samples in the limited-size memory buffer, while discard the rest.
Aiming at extracting the representative samples from non-\textit{iid} streams without supervision, \Method employs
the Part and Select Algorithm (PSA)~\cite{salomon2013psa} for uniform subset selection.
PSA first performs $M$ partition steps which divide all samples into $M$ subsets, then picks one sample from each subset. Each step partitions the existing set with the greatest dissimilarity among its members, thus PSA selects a subset of samples with uniform distribution in the spanned feature space. 
To the best of the authors' knowledge, this is the first time using uniform subset selection in lifelong learning problems.


\vspace{-2mm}
\section{Evaluation}
\label{sec:evaluations}

\subsection{Experimental Setup}

\begin{figure*}[t!]
    \centering
    \begin{subfigure}[t]{0.9\textwidth}
        \centering
        \includegraphics[height=0.18in]{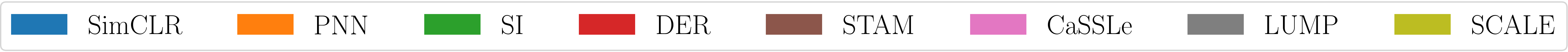}
    \end{subfigure}\\
    \begin{subfigure}[t]{0.18\textwidth}
        \centering
        \includegraphics[height=1.1in]{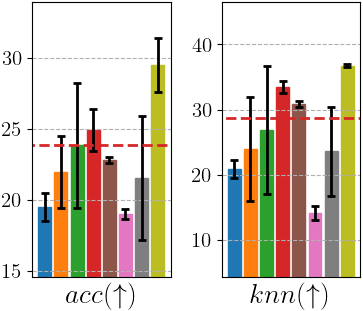}
        \vspace{-5mm}
        \caption{\small CIFAR-10 \textit{iid}}
    \end{subfigure}
    ~
    \begin{subfigure}[t]{0.18\textwidth}
        \centering
        \includegraphics[height=1.1in]{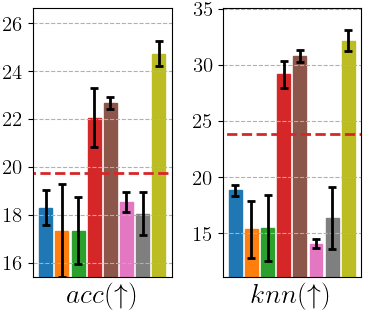}
        \vspace{-5mm}
        \caption{\small CIFAR-10 seq}
    \end{subfigure}
    ~
    \begin{subfigure}[t]{0.18\textwidth}
        \centering
        \includegraphics[height=1.1in]{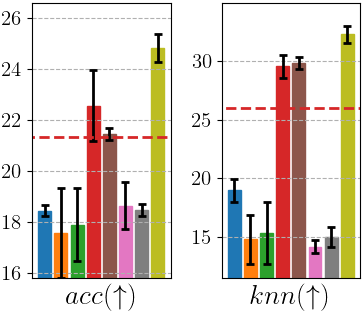}
        \vspace{-5mm}
        \caption{\small CIFAR-10 seq-bl}
    \end{subfigure}
    ~
    \begin{subfigure}[t]{0.18\textwidth}
        \centering
        \includegraphics[height=1.1in]{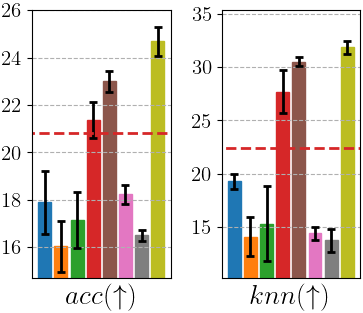}
        \vspace{-5mm}
        \caption{\small CIFAR-10 seq-im}
    \end{subfigure}
    ~
    \begin{subfigure}[t]{0.18\textwidth}
        \centering
        \includegraphics[height=1.1in]{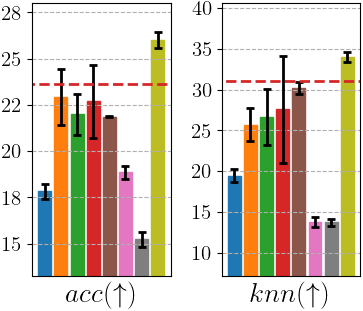}
        \vspace{-5mm}
        \caption{\small CIFAR-10 seq-cc}
    \end{subfigure}
    \begin{subfigure}[t]{0.18\textwidth}
        \centering
        \includegraphics[height=1.1in]{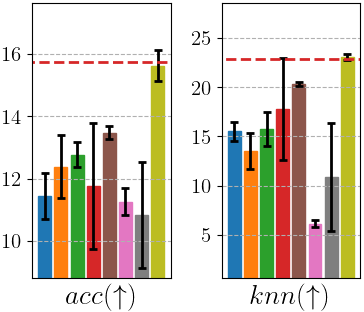}
        \vspace{-5mm}
        \caption{\small CIFAR-100 \textit{iid}}
    \end{subfigure}
    ~
    \begin{subfigure}[t]{0.18\textwidth}
        \centering
        \includegraphics[height=1.1in]{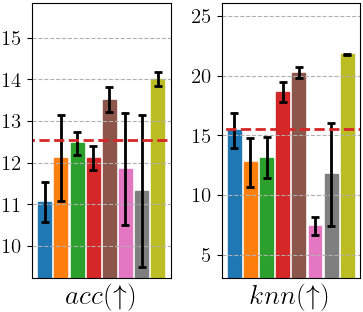}
        \vspace{-5mm}
        \caption{\small CIFAR-100 seq}
    \end{subfigure}
    ~
    \begin{subfigure}[t]{0.18\textwidth}
        \centering
        \includegraphics[height=1.1in]{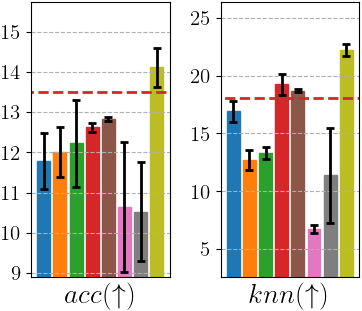}
        \vspace{-5mm}
        \caption{\small CIFAR-100 seq-bl}
    \end{subfigure}
    ~
    \begin{subfigure}[t]{0.18\textwidth}
        \centering
        \includegraphics[height=1.1in]{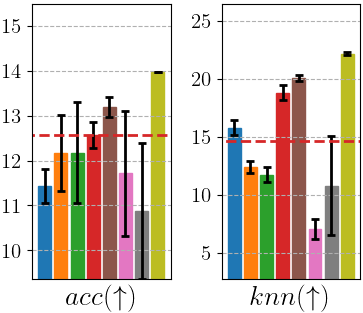}
        \vspace{-5mm}
        \caption{\small CIFAR-100 seq-im}
    \end{subfigure}
    ~
    \begin{subfigure}[t]{0.18\textwidth}
        \centering
        \includegraphics[height=1.1in]{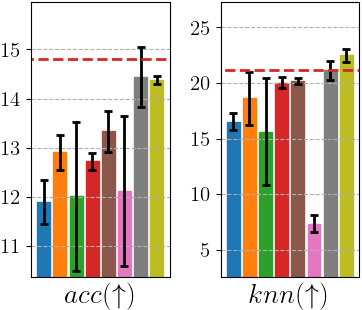}
        \vspace{-5mm}
        \caption{\small CIFAR-100 seq-cc}
    \end{subfigure} \\
    \begin{subfigure}[t]{0.18\textwidth}
        \centering
        \includegraphics[height=1.1in]{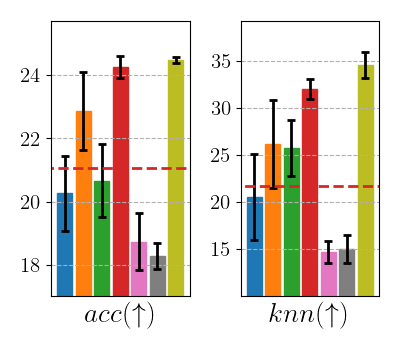}
        \vspace{-5mm}
        \caption{\small TinyImageNet \textit{iid}}
    \end{subfigure}
    ~
    \begin{subfigure}[t]{0.18\textwidth}
        \centering
        \includegraphics[height=1.1in]{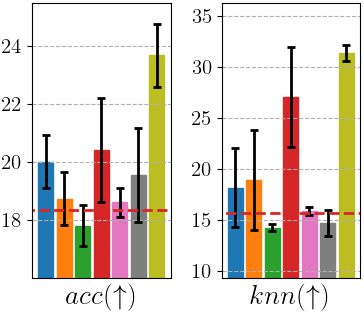}
        \vspace{-5mm}
        \caption{\small TinyImageNet seq}
    \end{subfigure}
    ~
    \begin{subfigure}[t]{0.18\textwidth}
        \centering
        \includegraphics[height=1.1in]{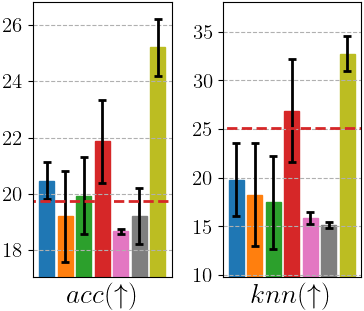}
        \vspace{-5mm}
        \caption{\small TinyImageNet seq-bl}
    \end{subfigure}
    ~
    \begin{subfigure}[t]{0.18\textwidth}
        \centering
        \includegraphics[height=1.1in]{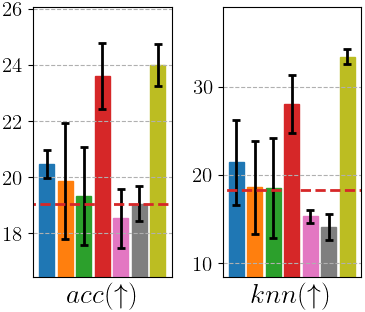}
        \vspace{-5mm}
        \caption{\small TinyImageNet seq-im}
    \end{subfigure}
    ~
    \begin{subfigure}[t]{0.18\textwidth}
        \centering
        \includegraphics[height=1.1in]{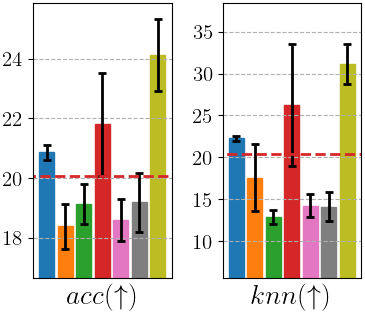}
        \vspace{-5mm}
        \caption{\small TinyImageNet seq-cc}
    \end{subfigure} \\
    \vspace{-2mm}
    \caption{\small \textbf{\Method improves $k$NN accuracy over the best state-of-the-art baseline by up to 3.83\%, 2.77\% and 5.86\% kNN on CIFAR-10, CIFAR-100 and TinyImageNet datasets.} The figures show final accuracy results on five different streams sampled from CIFAR-10 (first row), CIFAR-100 (second row) and TinyImageNet (third row) datasets. For each data stream setting, the left figure displays ACC while the right figure shows the $k$NN accuracy at the end of the stream. The \textcolor{red}{red} dashed line depicts the ACC or $k$NN accuracy of SupCon.}
    \vspace{-5mm}
    \label{fig:acc}
\end{figure*}


\noindent \textbf{Datasets:} We construct the online single-pass data streams from 
CIFAR-10 (10 classes)~\cite{Netzer2011}, CIFAR-100 (20 coarse classes)~\cite{krizhevsky2009learning} and a subset of TinyImageNet (10 classes)~\cite{deng2009imagenet}.
For each dataset, we construct five types of streams:
\textbf{\textit{iid}}, sequential classes (\textbf{seq}), sequential classes with blurred boundaries (\textbf{seq-bl}), sequential classes with imbalance lengths (\textbf{seq-im}), and sequential classes with concurrent classes (\textbf{seq-cc}).

\noindent \textbf{Networks:}
For all datasets, we apply ResNet-18~\cite{he2016deep} with a feature space dimension of 512.

\noindent \textbf{Baselines.}
Since \Method uses an InfoNCE-based loss, we compare with \textbf{SimCLR}~\cite{chen2020simple} and \textbf{SupCon}~\cite{khosla2020supervised} and the following lifelong learning baselines using SimCLR as backbone:
\vspace{-2mm}
\begin{itemize}[itemsep=-2mm]
    \item From the group of supervised lifelong learning, we select \textbf{PNN}~\cite{rusu2016progressive}, \textbf{SI}~\cite{zenke2017continual} and \textbf{DER}~\cite{buzzega2020dark} with necessary modifications for online ULL.
    \item For task-based ULL, we  use the source code of  \textbf{CaSSLe}~\cite{fini2021self} after removing the task labels.
    \item Finally, we also compare with \textbf{STAM}~\cite{smith2019unsupervised}, using their original data loader and parameters, and \textbf{LUMP}~\cite{madaan2022representational}.
\end{itemize}
\vspace{-2mm}
We did not compare with VAE-based methods such as~\cite{jiang2017variational,rao2019continual} since they are reported to scale poorly on medium to large image datasets~\cite{falcon2021aavae}.

\noindent \textbf{Metrics.} We use spectral clustering with $T$ as the number of clusters and compute the ACC. $k$NN classifier is used to evaluate the supervised accuracy with $k=50$.


Implementation details of \Method and baselines are presented in the Appendix.
All memory methods use a buffer of size $M=1280$. The size of the sampled memory batch is $m=128$, which is the same as the size of the streaming batch $n$.
For the similarity threshold, we use an adaptive threshold of $mean + \mu (max - mean)$ where $mean$ and $max$ are the mean and max pairwise similarity in $p_{ij}$. With an adaptive threshold, we alleviate the effects of variations in absolute similarity. 
\Method employs the Stochastic Gradient Descent optimizer with a learning rate of 0.03.



\subsection{Accuracy Results}
\label{sec:accuracy-results}
\noindent \textbf{Final Accuracy.} 
The final ACC and $k$NN accuracy on all datasets and all data streams are reported in Figure~\ref{fig:acc}.
Both mean and standard deviation of the accuracy are reported after 3 random trials.
ACC values are generally lower than their $k$NN counterparts. 
It should be noted that \Method outperforms all state-of-the-art ULL algorithms on almost all streaming patterns, both in terms of ACC and $k$NN accuracy.
In all settings in CIFAR-10, \Method improves 1.69-4.62\% on ACC and 1.32-3.83\% on $k$NN comparing with the best performed baseline. 
For CIFAR-100, \Method achieves improvements of up to 2.15\% regarding ACC and 2.77\% regarding $k$NN comparing with the best baseline.
For TinyImageNet, \Method enhances 0.2-3.33\% on ACC and 2.53-5.86\% on $k$NN accuracy over the best baseline.
Out of all data streams, \textit{iid} and seq-cc streams are easier to learn while the single-class sequential streams are more challenging and result in lower accuracy.
Our results demonstrate the strong adaptability of \Method which does not require any prior knowledge about the data stream. 

\noindent \textbf{Baseline Performances.} SimCLR produces low accuracy as it is originally designed for offline unsupervised representation learning with multiple epochs. Interestingly, the supervised contrastive learning baseline, SupCon (shown by red dashed line in Figure~\ref{fig:acc}), does not always result in superior accuracy and can be attributed to overfitting on the limited memory buffer.
Such result aligns with the recent findings that self-supervisedly learned representations are more robust than supervised counterparts under non-\textit{iid} streams~\cite{fini2021self,liu2021self}.
Among the techniques adapted from supervised lifelong learning, DER achieves relatively good results on all datasets but is still not comparable with \Method.
The recently proposed ULL module, CaSSLe , significantly relies on task boundary knowledge to preserve the classification semantics from previous tasks, thus showing poor results in our online ULL setup.
LUMP utilizes a mixup data augmentation technique and may not work well for certain image datasets.
STAM outperforms the rest of the baselines. However, STAM utilizes a unique memory architecture and cannot be fine-tuned for downstream tasks.

\noindent \textbf{Accuracy Curve.} To examine the dynamics of online learning, we summarize the $k$NN accuracy curves during training on blurred sequential CIFAR-10 and CIFAR-100 streams in Figure~\ref{fig:acc-curve-seq-knn} (more results in the Appendix). We can observe that \Method enjoys gradually increasing $k$NN accuracy as we introduce new classes, which demonstrates \Method's ability to consistently learn new knowledge while consolidating past information, all without supervision or prior knowledge. Most baselines are subject to collapse or forgetting, and are not able to distill or remember the knowledge in online ULL. The expandable-memory baseline STAM is incapable of learning without effective novelty detection. 

\begin{figure}[t!]
    \centering
    \begin{subfigure}[t]{0.5\textwidth}
        \centering
        \includegraphics[height=0.26in]{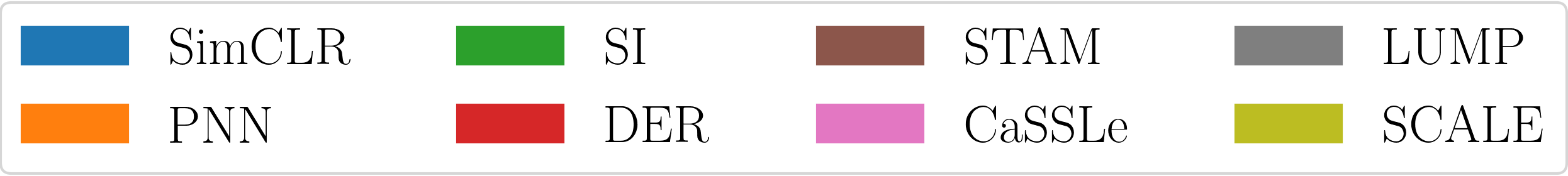}
    \end{subfigure}\\
    \begin{subfigure}[t]{0.48\textwidth}
        \centering
        \includegraphics[height=0.9in]{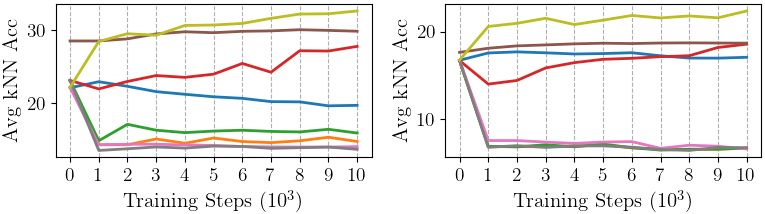}
    \end{subfigure} \\
    \vspace{-2mm}
    \caption{\small The average $k$NN accuracy during training on the \textit{blurred sequential} streams sampled from CIFAR-10 (left) and CIFAR-100 (right) datasets. Each training trial contains 10k training steps while each class spans 1k steps.}
    \vspace{-2mm}
    \label{fig:acc-curve-seq-knn}
\end{figure}

\begin{table}[!t]
\small
\centering
\caption{\small Average final $k$NN accuracy on the \textit{sequential} streams, under different combinations of loss functions.}
\vspace{-2mm}
\label{tbl:ablation-loss}
\begin{tabular}{c|c|c|c} 
\toprule
\small
Contrast Loss & Forget Loss & CIFAR-10 & TinyImageNet \\ \hline
SimCLR~\cite{chen2020simple} & \X & 18.84 & 18.13 \\ 
SupCon~\cite{khosla2020supervised} & \X & 23.83 & 15.67 \\
Co2L~\cite{cha2021co2l} & \C & 30.63 & 30.80 \\
{\Method} & \X & 30.45 & 30.40 \\
{\Method} & \C & \textbf{32.18} & \textbf{31.33} \\
\bottomrule
\end{tabular}
\end{table}


\subsection{Ablation Studies}
\vspace{-2mm}
\noindent \textbf{Loss Functions.}
We experiment with various combinations of contrastive loss and forgetting loss on the \textit{sequential} streams, as shown in Table~\ref{tbl:ablation-loss}.
Even with a replay buffer, SimCLR and SupCon do not lead to satisfying results on online ULL.
Co2L~\cite{cha2021co2l} is a supervised lifelong learning baseline using constrastive and forgetting losses. For fair comparison, we remove its dependence on class labels.
With the pseudo-supervised contrastive loss, SCALE gains 1.55\% in terms of $k$NN accuracy on CIFAR-10 compared to Co2L with a traditional contrastive loss.
With the forgetting loss, SCALE gets 1.78\% $k$NN accuracy gain on CIFAR-10. 

\noindent \textbf{Memory Update Policies.}
We experiment with \Method on the \textit{imbalanced sequential} stream with different memory update policies and summarize the results in Table~\ref{tbl:ablation-memory}.
With the distribution-aware uniform PSA memory update, SCALE surpasses the rest unsupervised strategies.
KMeans-based memory selection does not lead to the best result on sequential streams as the representations are not separable.
MinRed~\cite{purushwalkam2022challenges} prioritizes dissimilar samples regardless of global distribution, thus leads to biased selection and degraded performance on imbalanced data.
All components of SCALE are necessary for the best overall learning performance.

\begin{table}[!t]
\small
\centering
\vspace{-2mm}
\caption{\small Average final $k$NN accuracy on the \textit{imbalanced sequential} streams using different memory update policies in {\Method}.}
\vspace{-2mm}
\label{tbl:ablation-memory}
\begin{tabular}{c|c|c|c} 
\toprule
\small
Memory update & CIFAR-10 & CIFAR-100 & TinyImageNet \\ \hline
w/ label & 32.41 & 21.21 & 27.73 \\ \hline
random & 29.80 & 20.10 & 23.67 \\ 
KMeans & 31.59 & 22.15 & 29.07 \\
MinRed~\cite{purushwalkam2022challenges} & 23.66 & 19.75 & 25.13 \\
PSA (this paper) & \textbf{32.21} & \textbf{23.16} & \textbf{31.33} \\
\bottomrule
\end{tabular}
\end{table}


\subsection{Hyperparameters}

\begin{figure}[t!]
    \centering
    \begin{subfigure}[t]{0.48\textwidth}
        \centering
        \includegraphics[height=0.9in]{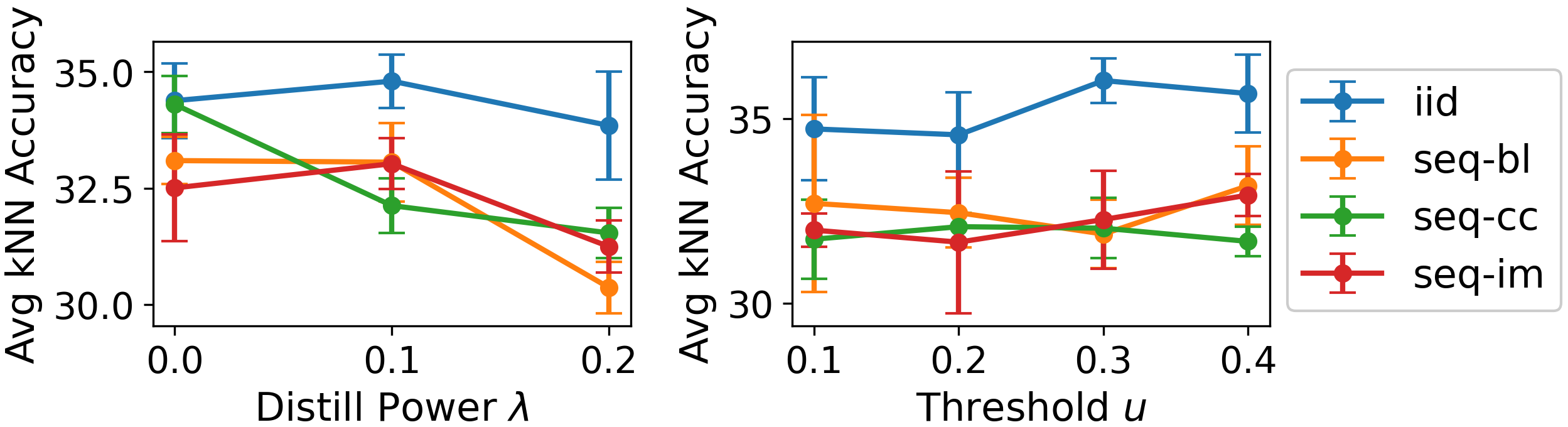}
    \end{subfigure} \\
    \vspace{-2mm}
    \caption{\small Experiments of hyperparameters on CIFAR-10 streams under various $\lambda$ (left) and $u$ (right).}
    \vspace{-4mm}
    \label{fig:sensitivity}
\end{figure}

We experiment with the important parameters in \Method. The weight balancing coefficient $\lambda$ plays an important role in the balance between pseudo-contrastive loss and self-supervised forgetting loss in \Method. The accuracy on various CIFAR-10 streams after 3 random trials, under various $\lambda$, are plotted in Figure~\ref{fig:sensitivity} (left). In most settings, $\lambda=0.1$ produces the best results.
A smaller $\lambda$ places less weight on the forgetting loss thus leads to forgetting; conversely, a larger $\lambda$ may over-emphasize the memorizing effect and prevent learning meaningful representations. 

The threshold $u$ is critical in defining the pseudo-positive set. The accuracy after 3 random trials on CIFAR-10 streams are shown in Figure~\ref{fig:sensitivity} (right). The sensitivity to threshold on \textit{iid} and sequential streams is different. For \textit{iid} streams, each incoming batch contains diverse samples from all classes. A higher threshold improves performance by restricting the pseudo-positive set to near-by samples that are more likely to belong to one class. For sequential streams, as the samples from the same batch are from the same class, a positive but lower threshold helps filter sufficiently similar samples into the pseudo-positive set, to boost learning outcome. 

\section{Conclusion}
\label{sec:conclusions}
Existing works in unsupervised lifelong learning assume various prior knowledge thus are not applicable for learning in the wild.
In this paper, we propose the online unsupervised lifelong learning problem without prior knowledge that (i) accepts non-\textit{iid}, non-stationary and single-pass streams, (ii) does not rely on external supervision, and (iii) does not assume prior knowledge. 
We propose \Method, a self-supervised contrastive lifelong learning technique based on pairwise similarity. \Method uses a pseudo-supervised contrastive loss for representation learning, a self-supervised forgetting loss to avoid catastrophic forgetting, and an online memory update for uniform subset selection. 
Experiments demonstrate that \Method improves $k$NN accuracy over the best state-of-the-art baseline by up to 3.83\%, 2.77\% and 5.86\% on all non-\textit{iid} CIFAR-10, CIFAR-100 and TinyImageNet streams.

\section*{Acknowledgements}
This work was supported in part by National Science Foundation under Grants \#2003279, \#1830399, \#1826967, \#2100237, \#2112167, \#2112665, and in part by SRC under task \#3021.001. This work was also supported by a startup funding by the University of Texas at Dallas.

{\small
\bibliographystyle{ieee_fullname}
\bibliography{egbib}
}

\section*{Appendix}
\appendix
\input{supplementary}

\end{document}

%% file: supplementary.tex
In the supplementary, we include more details on the following aspects:
\begin{itemize}[itemsep=-1.5mm]
    \item In Section~\ref{sec:implementation}, we list the \textbf{implementation details of \Method, lifelong learning baselines and self-supervised learning baselines}, especially the hyperparameters for each dataset. For \Method, we detail the online memory update algorithm and compare with MinRed~\cite{purushwalkam2022challenges}.
    \item In Section~\ref{sec:data-stream}, we provide details on \textbf{constructing data streams} in our online unsupervised lifelong learning problem setup.
    \item In Section~\ref{sec:acc-curve}, we show the \textbf{accuracy curves} during training on all datasets. The accuracy curves of all lifelong learning baselines and \Method are complementary to the results in Section 6 of the main paper.
    \item In Section~\ref{sec:batch-size}, we conduct sensitivity analyses on the \textbf{streaming batch size $n$ and memory batch size $m$} in \Method.
    \item In Section~\ref{sec:temperature}, we conduct sensitivity analyses on the \textbf{temperature $\tau$} in our pseudo-supervised contrastive loss. Different temperatures are ideal for \textit{iid} and non\textit{iid} streams.
    \item In Section~\ref{sec:tsne}, we present the \textbf{t-SNE plots} of the features during periodic evaluation, which vividly demonstrates \Method's learning process.
    \item In Section~\ref{sec:compute-complexity}, we analyze the \textbf{computation time complexity} of \Method, including all components of pseudo-contrastive loss, forgetting loss and memory update.
\end{itemize}

\section{Implementation Details}
\label{sec:implementation}

\subsection{Implementation Details of \Method}
\label{sec:implement-scale}
We implement the pseudo-supervised contrastive learning component of \Method based on the official SupCon framework~\cite{khosla2020supervised}. We use ResNet-18~\cite{he2016deep} with a feature space dimension of 512 as backbone. We use the Stochastic Gradient Descent (SGD) optimizer with learning rate of 0.03. The hyperparameters across all datasets are summarized in Table~\ref{tbl:hyperparameter}.

\begin{table}[th]
\small
\centering
\vspace{-2mm}
\caption{\small Hyperparameters of SCALE across all datasets.}
\vspace{-2mm}
\label{tbl:hyperparameter}
\begin{tabular}{c|c|c} 
\toprule
\small
Param. & Explain. & Value \\ \hline
$lr$ & Learning rate & 0.01 \\ 
$n$ & Batch size for streaming data & 128 \\
$M$ & Memory buffer size & 1280 \\
$m$ & Sampled memory batch size & 128 \\
$\tau$ & Temperature for pseudo-contrastive loss & 0.1 \\
$\mu$ & Relative similarity threshold & 0.05 \\
$\lambda$ & Weight for self-supervised forgetting loss & 0.1 \\
\bottomrule
\end{tabular}
\vspace{-2mm}
\end{table}

\textbf{Data augmentation.}
All methods except STAM share the same augmentation procedure.
For STAM, we use their official data loader with custom pre-processing.
During the training phase, our data augmentation procedure first normalizes the data using mean and variances.
we apply random scaling 0.2-1, random horizontal flip, random color jitter of brightness 0.6-1.4, contrast 0.6-1.4, saturation  0.6-1.4, hue 0.9-1.1, and random gray scale with $p=0.2$ for CIFAR-10 and CIFAR-100. For TinyImageNet, we apply the random scaling 0.08-1 with random aspect ratio 0.75-1.33 and 
bicubic interpolation. All images are resized to 32 $\times$ 32. 
During the evaluation phase, we only normalize the data but do not use any augmentation for all datasets.

\textbf{Uniform memory subset sampling.}
In SCALE, one key component is the online memory update where we adapt the uniform subset sampling algorithms. To achieve the best performance in online unsupervised lifelong learning (ULL), the memory buffer is supposed to retain the most ``representative'' samples regarding the historical distribution in the feature space. Uniform sampling mechanism is desired to extract representative samples from the sequential imbalanced streams. To remind the readers, the input to the memory update is the imbalanced combined features $\left \{ \mathbf{z}_i \right \}_{i=1}^{M+n}$ projected by the latest model $\theta_u^t$ from the previous raw memory samples $\left \{ \mathbf{e}_i\right \}_{i=1}^{M}$ and latest streaming batch $X_u^t$.
The goal is to select a subset of $M$ samples from $\left \{ \mathbf{z}_i \right \}_{i=1}^{M+n}$ then store the corresponding raw samples in the new memory buffer.
We employ the Part and Selection Algorithm (PSA)~\cite{salomon2013psa} in \Method and adapt the implementation from \texttt{diversipy} (\url{https://github.com/DavidWalz/diversipy}).
The implementation is the slightly improved version from~\cite{wessing2015two}. 
PSA is a linear-time algorithm designed to select a subset of well-spread points. The algorithm has two stages: first, the candidate set $\left \{ \mathbf{z}_i \right \}_{i=1}^{M+n}$ is partitioned into $M$ subsets, then one member from each subset is selected to form the updated memory.
During the first stage, each partition step selects the set with the greatest dissimilarity among its members to divide. The dissimilarity of a set $A=\left \{ \mathbf{z}_i \right \}_{i=1}^{M+n}$ is defined as the maximum absolute difference among all dimensions:
\begin{subequations}
\begin{align}
    & a_j := \min_{i=1,...,M+n} z_{ij}, \quad b_j:=\max_{i=1,...,M+N} z_{ij}, \notag \\
    & \Delta_j = b_j - a_j, \quad j = 1,...,K \\
    & \o A := \max_{j=1,...,K} \Delta_j
\end{align}
\end{subequations}
where $K$ denotes the dimension of the feature space $\mathcal{Z}$. The dissimilarity of $A$ is the diameter of $A$ in the Chebyshev metric. During the second stage, PSA chooses the closest member (in Euclidean metric) to the center of the hyperrectangle around $A_i$.
The pseudocode and complexity analysis of PSA are presented in~\cite{salomon2013psa} and~\cite{wessing2015two}. The execution time of PSA in our setup is discussed in Section~\ref{sec:compute-complexity}.

\textbf{Comparison of MinRed and PSA (in \Method).}
The latest study by Purushwalkam \textit{et al.}~\cite{purushwalkam2022challenges} proposed a minimum redundancy (MinRed) memory update policy, which assists buffer replay in self-supervised learning. When the number of samples in the memory exceeds its capacity, they rely on the cosine distance between all pairs of samples to discard the most redundant one:
\begin{equation}
    i^* = \argmin_i \min_{j \neq i} d_{cos}(\mathbf{z}_i, \mathbf{z}_j)
\end{equation}

\noindent Intuitively, MinRed is a greedy heuristic that keeps the most ``disimilar'' $M$ samples. The ``dissimilarity'' is judged by the greatest distance from its closest selected feature. 
Although MinRed is effective in retaining diverse samples, it does not take into account global distribution and may lead to biased selection on imbalanced incoming streams. This leads to a degraded performance, as shown in Table 3 of the main paper.

\subsection{Implementation Details of Lifelong Learning Baselines}
\label{sec:implement-lifelong}
The following lifelong learning baselines are used to compare with SCALE:
\begin{itemize}[itemsep=-1.5mm]
    \item \textbf{PNN}~\cite{rusu2016progressive}: Progressive Neural Network gradually expands the network architecture.
    \item \textbf{SI}~\cite{zenke2017continual}: Synaptic Intelligence performs online per-synapse consolidation as a typical regularization technique. 
    \item \textbf{DER}~\cite{buzzega2020dark}: Dark Experience Replay retains existing knowledge by matching the network logits across a sequence of tasks.
    \item \textbf{STAM}~\cite{smith2019unsupervised} uses online clustering and novelty detection to update an expandable memory architecture. 
    \item \textbf{CaSSLe}~\cite{fini2021self} proposes a general framework that extracts the best possible representations invariant to task shifts in ULL. 
    \item \textbf{LUMP}~\cite{madaan2022representational} interpolates the current with the previous samples to alleviate catastrophic forgetting in ULL. Use SimCLR.
\end{itemize}
Note, that all methods except STAM are addable to self-supervised learning backbones, while STAM employs a unique expandable memory architecture. As \Method lies on the SimCLR backbone, we also experiment with the above baselines on the SimCLR backbone for a fair comparison. 
We did not compare with VAE-based methods such as~\cite{jiang2017variational,rao2019continual} since they have been reported to scale poorly on large image datasets~\cite{falcon2021aavae}.
More implementation details are grouped and summarized as follows:
\begin{itemize}[itemsep=-1.5mm]

    \item \textbf{PNN}, \textbf{SI}, \textbf{DER}, \textbf{LUMP} are adapted from the official framework in~\cite{madaan2022representational} using their default hyperparameters. PNN, SI and DER are originally designed for supervised lifelong learning but are adapted to ULL tasks as described in~\cite{madaan2022representational}. For fair comparison, we use SimCLR as the underlying contrastive learning backbone for these baselines. For DER and LUMP, we use a replay buffer of the same size as SCALE.
    
    \item We take advantage of the official implementation of \textbf{STAM} on CIFAR-10 and CIFAR-100 with their default hyperparameters.
    We use the original data loader and parameters for CIFAR-10, CIFAR-100 as in the released code, and use our clustering and $k$NN classifier on the learned embeddings.
    
    \item We use a modified version \textbf{CaSSLe} based on the original implementation. Specifically, we remove task labels and force the model to compare the representations of the current and previous batch.
\end{itemize}

\section{Data Streams Construction}
\label{sec:data-stream}
To remind the reader, we evaluated three image datasets: CIFAR-10 (10 classes)~\cite{Netzer2011}, CIFAR-100 (20 coarse classes)~\cite{krizhevsky2009learning} and a subset of ImageNet (10 classes)~\cite{deng2009imagenet}.
We construct five single-pass data streams for training:
\begin{itemize}[itemsep=-1.5mm]
    \item \textbf{\textit{iid} stream:} We sample 4096, 2560 and 500 images from each class of CIFAR-10, CIFAR-100, and TinyImageNet, then shuffle all samples.
    \item \textbf{Sequential class-incremental stream:} We sample 4096, 2560 and 500 images from each class of CIFAR-10, CIFAR-100, and TinyImageNet, then feed them class-by-class to the model.
    \item \textbf{Sequential class-incremental stream with blurred boundaries:} We sample the same number of images from each class as the standard sequential class-incremental stream. We then mix the last 25\% samples of the previous class with the first 25\% samples of the next class, with a gradual mix probability between 0.05 and 0.5. Specifically, for samples closer to the boundary, there is a higher probability to be exchanged with a sample on the other side of the boundary.
    \item \textbf{Sequential class-incremental stream with imbalanced class appearance:} For each incrementally introduced class, we randomly sample a subset with more than half of the total samples in that class. Specifically, suppose that there are $U$ samples in that class. We first uniformly sample an integer $V \in [0.5U, U]$, then we randomly sample $V$ samples from that class.
    \item \textbf{Sequential class-incremental stream with concurrent class appearance:} Similar as the sequential class-incremental stream, we sample the same amount of images from each class. We then group the classes 2-by-2 with its adjacent class, and shuffle all samples in one group. In this way, each 2-class group is revealed to the model incrementally, while the samples in one group follow a random order.
\end{itemize}
For the evaluation dataset, we sample 500, 250 and 50 samples per class from the official validation dataset of CIFAR-10, CIFAR-100 and TinyImageNet respectively.

\section{Accuracy Curve during Training}
\label{sec:acc-curve}

\begin{figure}[t]
    \centering
    \begin{subfigure}[t]{0.5\textwidth}
        \centering
        \includegraphics[height=0.25in]{figs/legend2.png}
    \end{subfigure}\\
    \begin{subfigure}[t]{0.45\textwidth}
        \centering
        \includegraphics[height=0.9in]{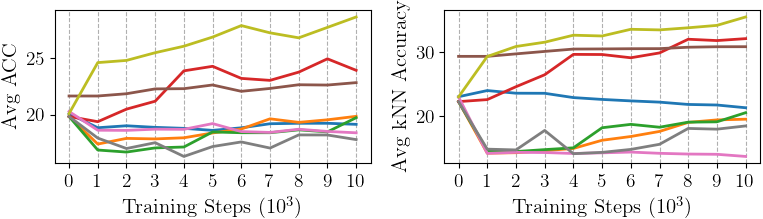}
        \caption{\small CIFAR-10 iid}
    \end{subfigure}\\
    \begin{subfigure}[t]{0.45\textwidth}
        \centering
        \includegraphics[height=0.9in]{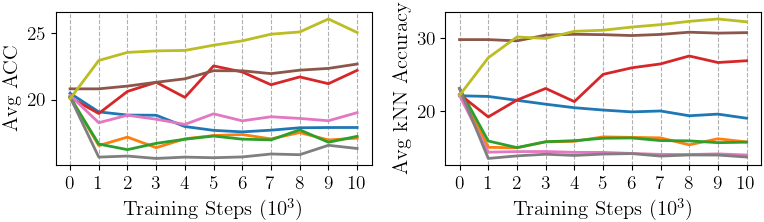}
        \caption{\small CIFAR-10 seq}
    \end{subfigure} \\
    \begin{subfigure}[t]{0.45\textwidth}
        \centering
        \includegraphics[height=0.9in]{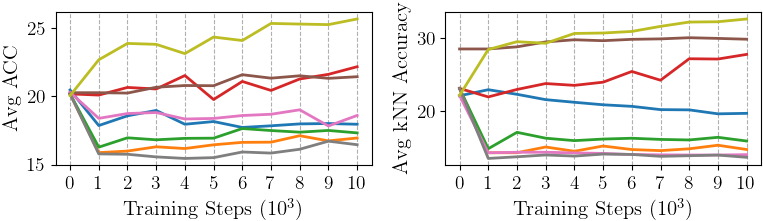}
        \caption{\small CIFAR-10 seq-bl}
    \end{subfigure}\\
    \begin{subfigure}[t]{0.45\textwidth}
        \centering
        \includegraphics[height=0.9in]{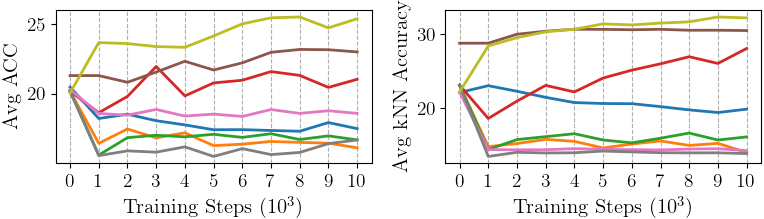}
        \caption{\small CIFAR-10 seq-im}
    \end{subfigure} \\
    \begin{subfigure}[t]{0.45\textwidth}
        \centering
        \includegraphics[height=0.9in]{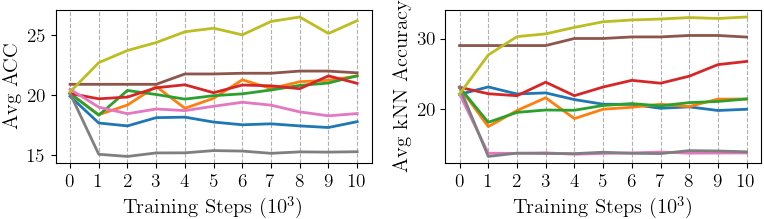}
        \caption{\small CIFAR-10 seq-cc}
    \end{subfigure}\\
    \vspace{-2mm}
    \caption{\small ACC and $k$NN accuracy curve on all streams sampled from CIFAR-10 using various lifelong learning baselines.}
    \vspace{-6mm}
    \label{fig:acc-curves-cifar-10}
\end{figure}

\begin{figure}[t]
    \centering
    \begin{subfigure}[t]{0.5\textwidth}
        \centering
        \includegraphics[height=0.25in]{figs/legend2.png}
    \end{subfigure}\\
    \begin{subfigure}[t]{0.45\textwidth}
        \centering
        \includegraphics[height=0.9in]{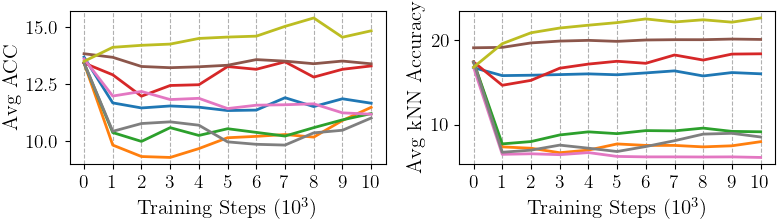}
        \caption{\small CIFAR-100 iid}
    \end{subfigure}\\
    \begin{subfigure}[t]{0.45\textwidth}
        \centering
        \includegraphics[height=0.9in]{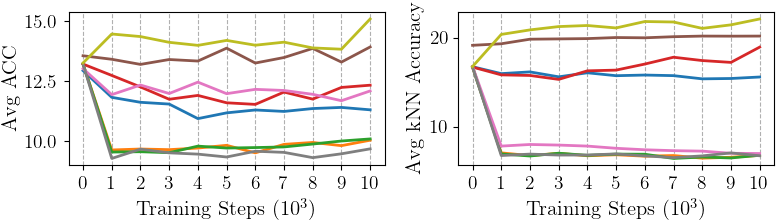}
        \caption{\small CIFAR-100 seq}
    \end{subfigure} \\
    \begin{subfigure}[t]{0.45\textwidth}
        \centering
        \includegraphics[height=0.9in]{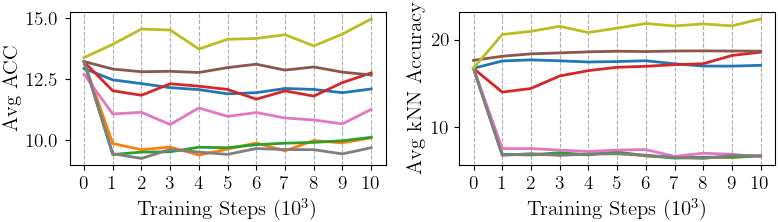}
        \caption{\small CIFAR-100 seq-bl}
    \end{subfigure}\\
    \begin{subfigure}[t]{0.45\textwidth}
        \centering
        \includegraphics[height=0.9in]{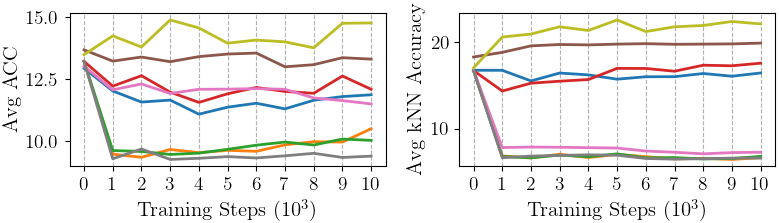}
        \caption{\small CIFAR-100 seq-im}
    \end{subfigure} \\
    \begin{subfigure}[t]{0.45\textwidth}
        \centering
        \includegraphics[height=0.9in]{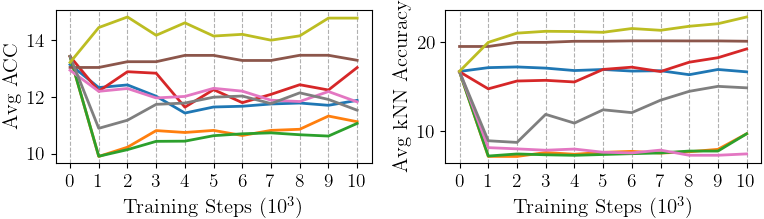}
        \caption{\small CIFAR-100 seq-cc}
    \end{subfigure}\\
    \vspace{-2mm}
    \caption{\small ACC and $k$NN accuracy curve on all streams sampled from CIFAR-100 using various lifelong learning baselines.}
    \vspace{-6mm}
    \label{fig:acc-curves-cifar-100}
\end{figure}

\begin{figure}[tbh]
    \centering
    \begin{subfigure}[t]{0.5\textwidth}
        \centering
        \includegraphics[height=0.25in]{figs/legend2.png}
    \end{subfigure}\\
    \begin{subfigure}[t]{0.45\textwidth}
        \centering
        \includegraphics[height=0.9in]{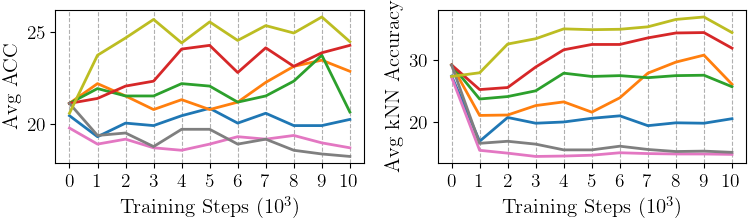}
        \caption{\small TinyImageNet iid}
    \end{subfigure}\\
    \begin{subfigure}[t]{0.45\textwidth}
        \centering
        \includegraphics[height=0.9in]{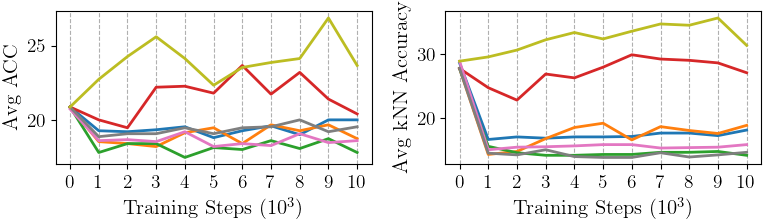}
        \caption{\small TinyImageNet seq}
    \end{subfigure} \\
    \begin{subfigure}[t]{0.45\textwidth}
        \centering
        \includegraphics[height=0.9in]{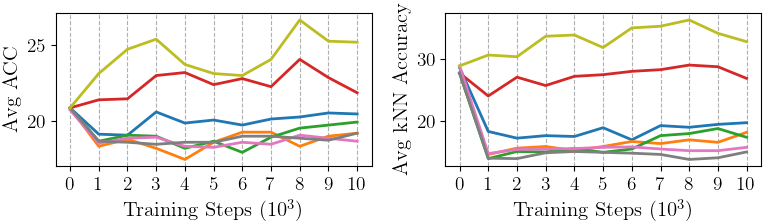}
        \caption{\small TinyImageNet seq-bl}
    \end{subfigure}\\
    \begin{subfigure}[t]{0.45\textwidth}
        \centering
        \includegraphics[height=0.9in]{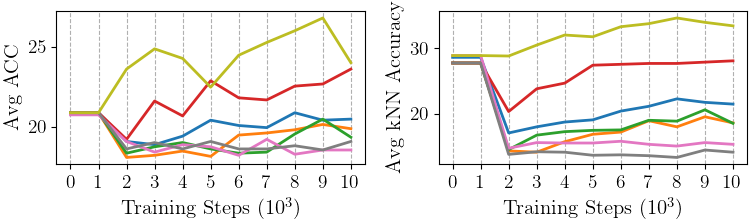}
        \caption{\small TinyImageNet seq-im}
    \end{subfigure} \\
    \begin{subfigure}[t]{0.45\textwidth}
        \centering
        \includegraphics[height=0.9in]{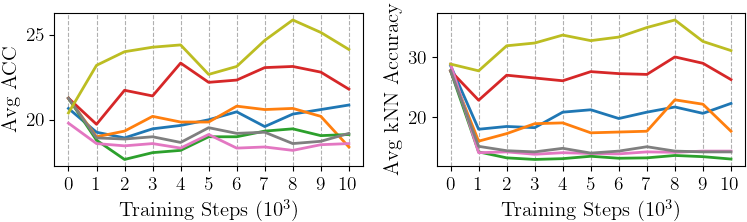}
        \caption{\small TinyImageNet seq-cc}
    \end{subfigure}\\
    \vspace{-2mm}
    \caption{\small ACC and $k$NN accuracy curve on all streams sampled from TinyImageNet using various lifelong learning baselines.}
    \vspace{-6mm}
    \label{fig:acc-curves-tinyimagenet}
\end{figure}

The accuracy curves of all lifelong learning methods during training are depicted in Figure~\ref{fig:acc-curves-cifar-10}, \ref{fig:acc-curves-cifar-100} and \ref{fig:acc-curves-tinyimagenet} for CIFAR-10, CIFAR-100 and TinyImageNet respectively. Outstanding from all methods, \Method learns incrementally regardless of the \textit{iid} or sequential manner. Compared to \textit{iid} cases, sequential data streams are more challenging, where more baselines present the ``forgetting'' or unimproved trend as new classes arrive. Among the three datasets, CIFAR-10 streams are easier to learn from. CIFAR-100 streams with 20 coarse classes act as the most challenging dataset where multiple baselines collapse from the beginning. The 10-class subset from ImageNet causes more fluctuations during the online learning procedure.



\section{Sensitivity Analyses of Streaming and Memory Batch Sizes} 
\label{sec:batch-size}
\begin{figure}[t]
\begin{center}
\centerline{\includegraphics[width=0.45\textwidth]{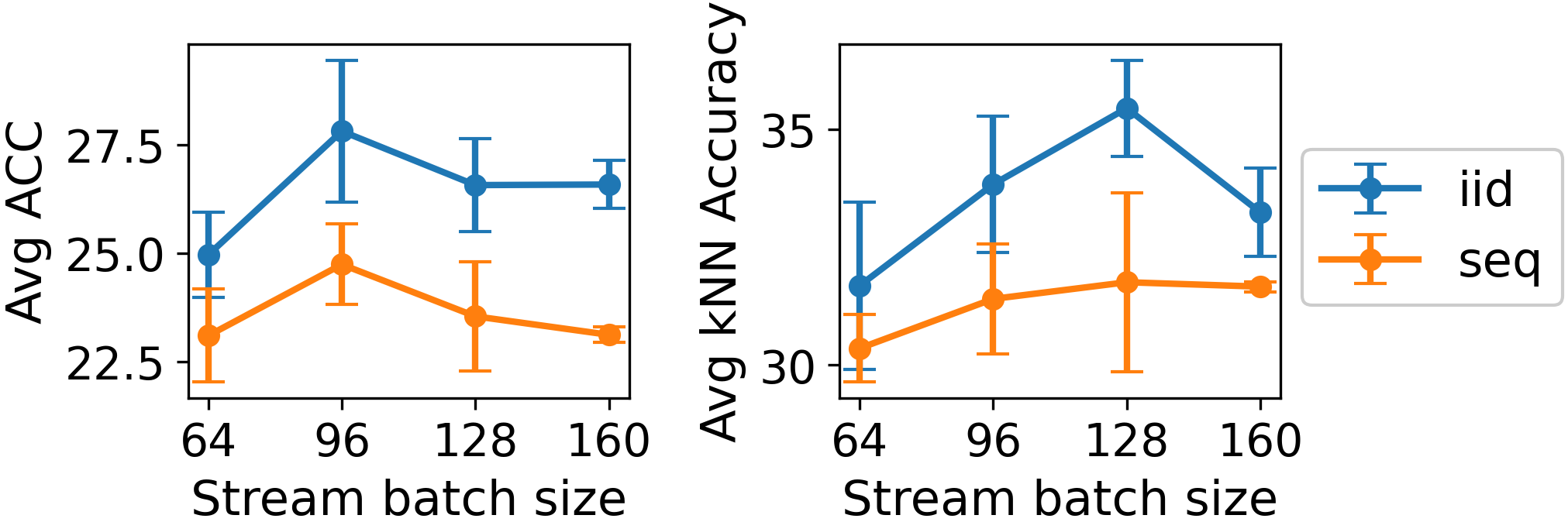}}
\vspace{-2mm}
\caption{\small Average ACC (left) and $k$NN accuracy (right) on \textit{iid} and sequential CIFAR-10 streams, with different batch sizes $n$ and memory batch size $m=128$.}
\label{fig:batch-size}
\end{center}
\vspace{-4mm}
\end{figure}

As indicated in multiple studies~\cite{chen2021exploring,grill2020bootstrap,zbontar2021barlow}, batch size has a significant impact on the performance of contrastive learning methods, as a large number of samples are required to enhance the contrast effect. We study the impact of streaming and memory batch sizes in \Method.
We first fix the memory batch size $m=128$ and alter the streaming batch size upon \textit{iid} and sequential CIFAR-10 streams. The average final ACC and $k$NN accuracy after 3 random trials are shown in Figure~\ref{fig:batch-size}. It can be seen that the impact of batch sizes on ACC and $k$NN accuracies is slightly different. 
Compared to ACC, $k$NN accuracy behaves more stably hence our discussion in the rest of the material mainly focuses on $k$NN accuracies.
For \textit{iid} streams, a larger batch size leads to a higher $k$NN accuracy in \Method, as more samples can be used for contrast. However, in the sequential case, \Method is robust to batch sizes with less than 1\% difference in terms of $k$NN accuracy when using batch sizes of 64, 96, 128 and 160.
Such robustness can be attributed to two reasons: (i) unlike SimCLR, we use small batch sizes for the online learning scenarios, thus the effect of varying batch sizes diminishes; (ii) for the sequential streams, the contrasting samples mainly come from the memory buffer (with different labels). Therefore a large batch size does not greatly improve the contrastive learning performance.

\begin{figure}[t]
\begin{center}
\centerline{\includegraphics[width=0.45\textwidth]{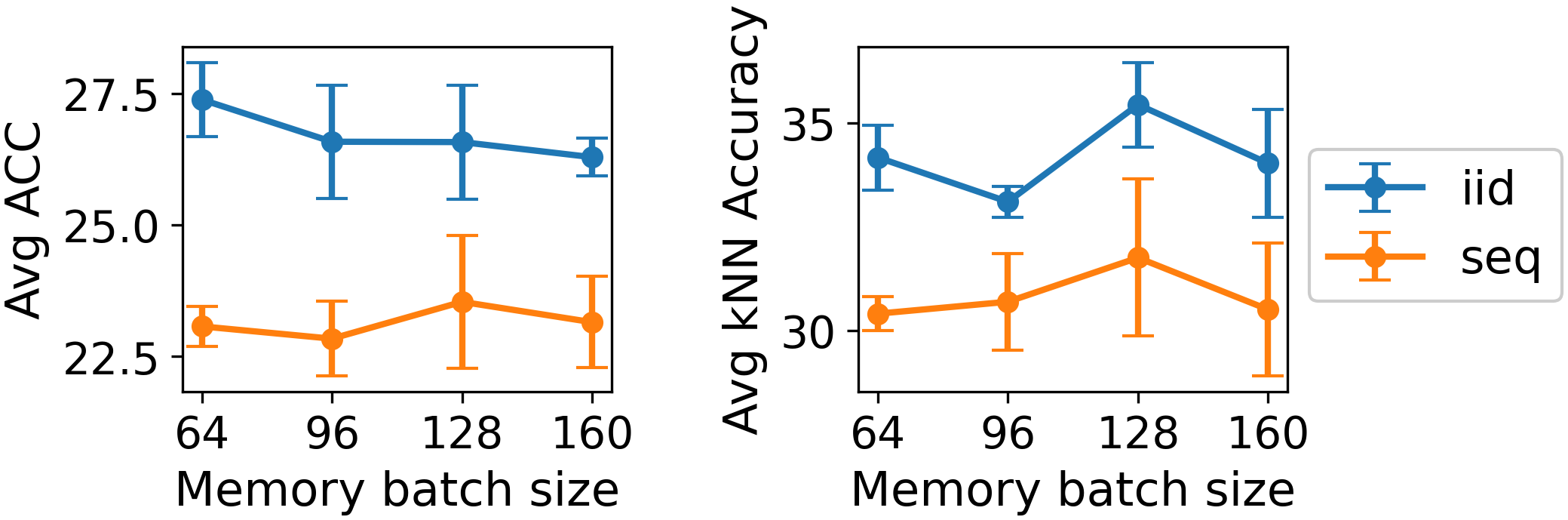}}
\vspace{-2mm}
\caption{\small Average ACC (left) and $k$NN accuracy (right) on \textit{iid} and sequential CIFAR-10 streams, with different memory batch sizes $m$ and streaming batch size $n=128$.}
\label{fig:mem-size}
\end{center}
\vspace{-4mm}
\end{figure}

We then fix the streaming batch size to $n=128$ and apply various memory batch sizes. The average ACC and $k$NN accuracy of \Method on 3 random CIFAR-10 streams is shown in Figure~\ref{fig:mem-size}. Interestingly, as the contrasting performance of \Method depends on both the streaming and memory samples, the effect of changing one of them is not significant. When using memory samples of 64, 96, 128 and 160 on sequential streams, the different on ACC and $k$NN accuracies are less than 0.7\% and 1.35\% respectively.

\section{Sensitivity Analyses of Temperature $\tau$} 
\label{sec:temperature}

\begin{figure}[t!]
\centering
\begin{subfigure}{0.22\textwidth}
    \centering
    \includegraphics[height=1.25in]{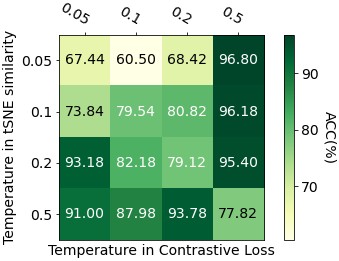}
\end{subfigure}
~~
\begin{subfigure}{0.22\textwidth}
    \centering        
    \includegraphics[height=1.25in]{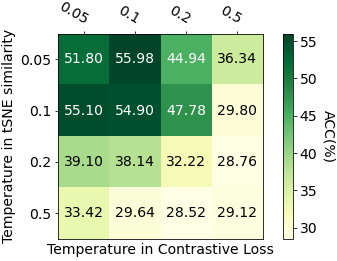}
\end{subfigure}
\vspace{-2mm}
\caption{\small Heatmap of final ACC on MNIST, \textit{iid} stream (left) and sequential class-incremental stream (right) using various temperatures.}
\vspace{-2mm}
\label{fig:temperature}
\end{figure}

We setup MNIST following similar protocols in Section~\ref{sec:data-stream}.
Figure~\ref{fig:temperature} reports the ACC at the end of \textit{iid} and single-class sequential data streams on MNIST, when choosing various values for temperature $\tau$ in the contrastive loss (Equation~(3)) and temperature $\kappa$ in the tSNE pseudo-positive set selection (Equation~(6)). It can be observed that the type of data stream (i.e., \textit{iid} or sequential) has a significant effect on the best combinations of temperatures. Under the \textit{iid} datastream, high temperature of $\tau=0.5$ is preferred while $\kappa$ has a small impact on the final ACC. However, in the sequential case, temperature of $\tau=0.1$ or even smaller is desired while $\kappa$ in pseudo-positive set construction also drives the final ACC. Intuitively, contrastive learning benefits when there are more negative samples from the other classes to compare against, where a large temperature value works better. However, in online ULL scenarios, a lower temperature $\tau$ with comparable $\kappa$ shows better performance in driving the closer samples together and memorizing the similarity relationship.

\section{t-SNE Plots during Training}
\label{sec:tsne}
\begin{figure}[t]
    \centering
    \begin{subfigure}[t]{0.15\textwidth}
        \centering
        \includegraphics[height=0.9in]{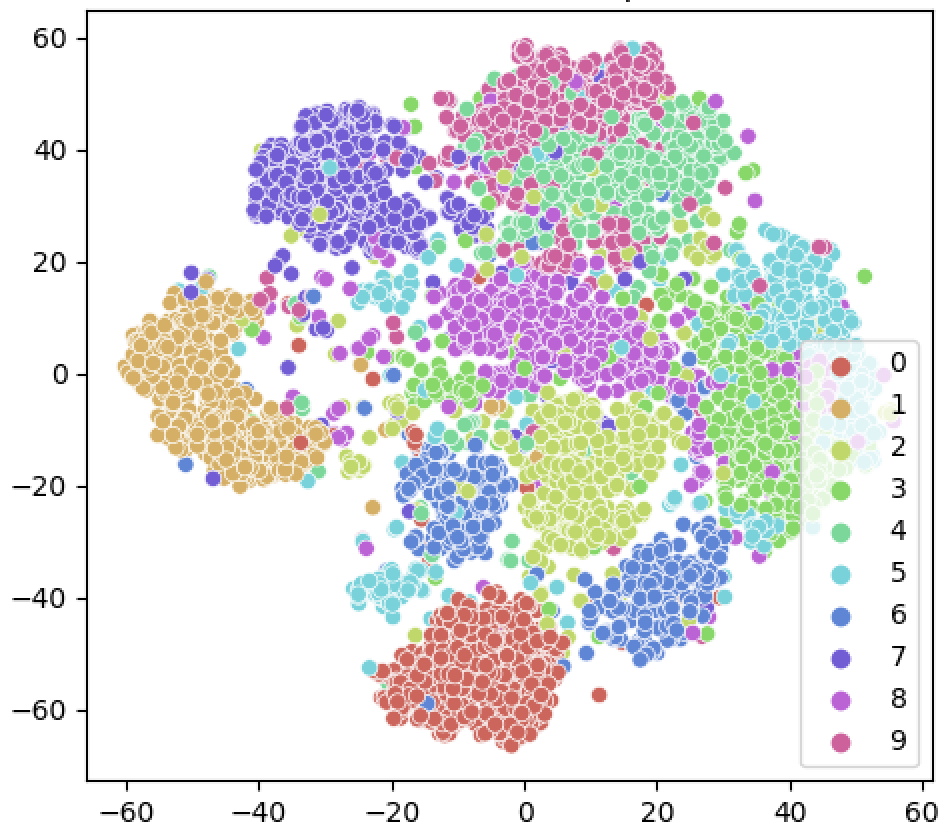}
    \end{subfigure}
    \begin{subfigure}[t]{0.15\textwidth}
        \centering
        \includegraphics[height=0.9in]{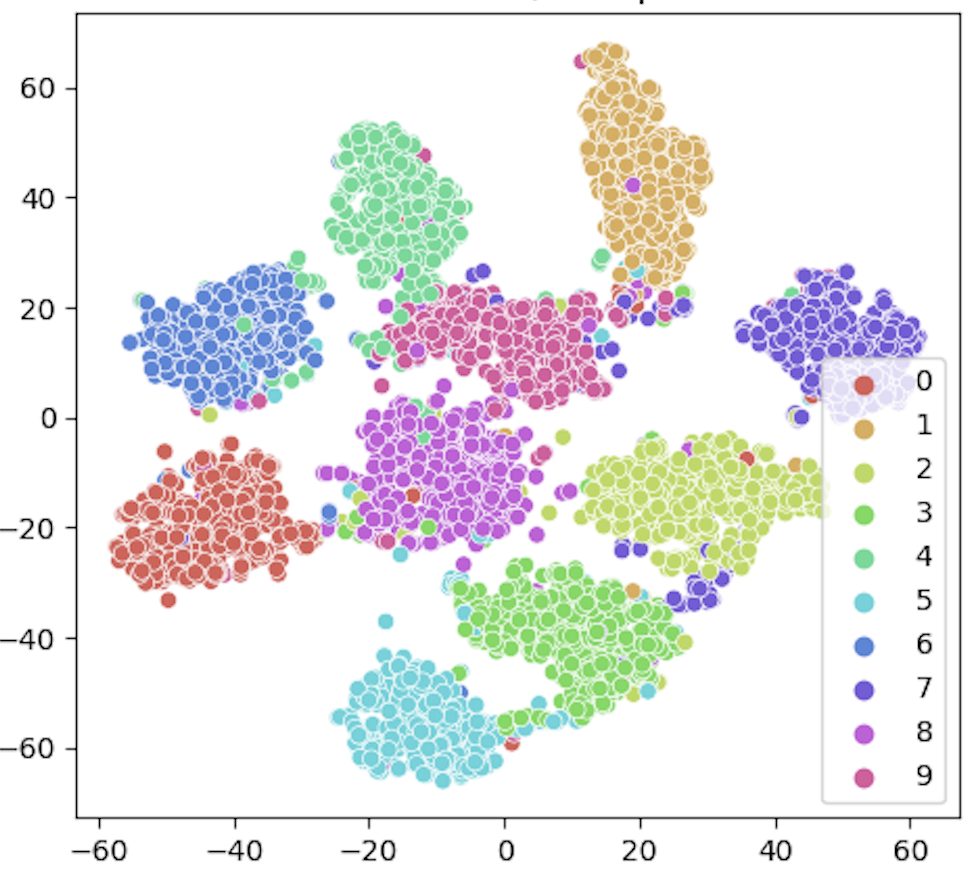}
    \end{subfigure}
    \begin{subfigure}[t]{0.15\textwidth}
        \centering
        \includegraphics[height=0.9in]{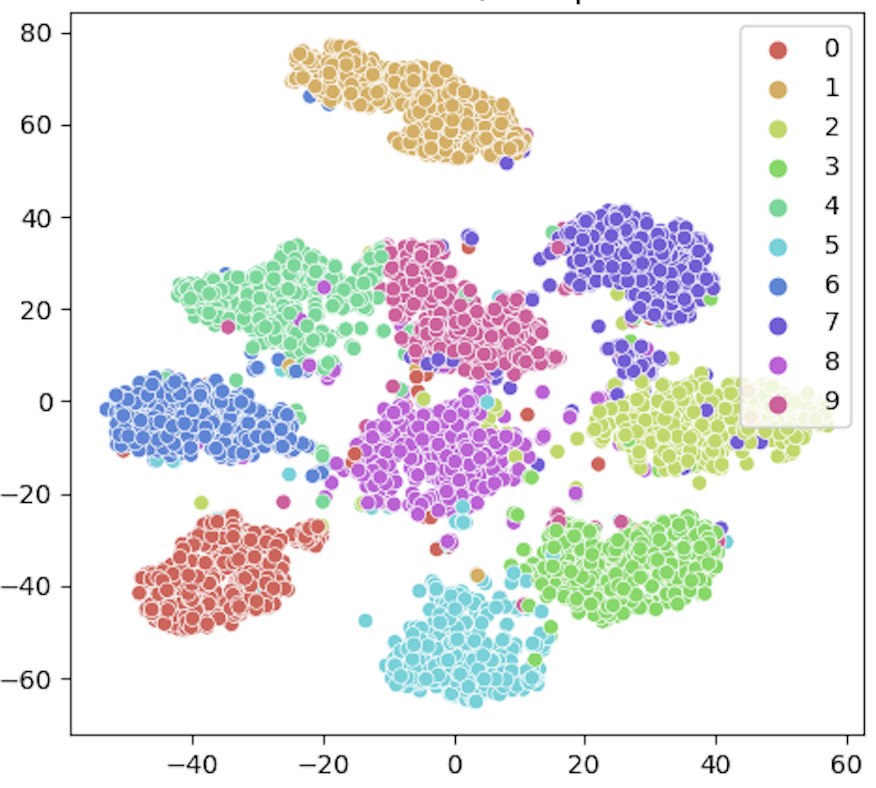}
    \end{subfigure}
    \vspace{-2mm}
    \caption{\small t-SNE plots on the evaluation dataset at the start (left), middle (middle) and end (right) of training on \textit{iid} data streams on MNIST.}
    \vspace{-4mm}
    \label{fig:tsne-iid}
\end{figure}

\begin{figure}[t]
    \centering
    \begin{subfigure}[t]{0.15\textwidth}
        \centering
        \includegraphics[height=0.9in]{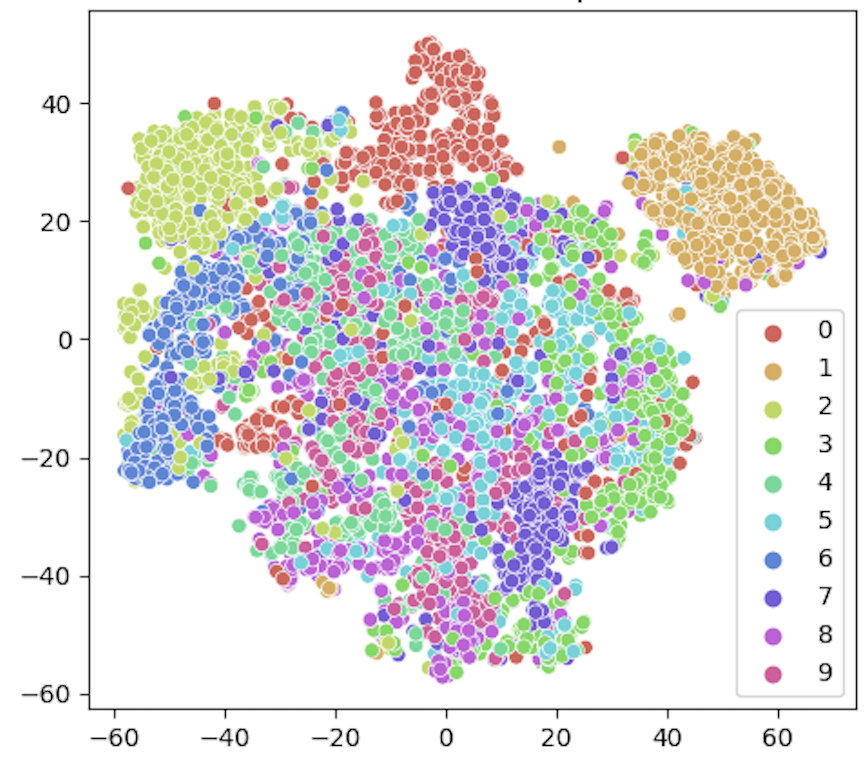}
    \end{subfigure}
    \begin{subfigure}[t]{0.15\textwidth}
        \centering
        \includegraphics[height=0.9in]{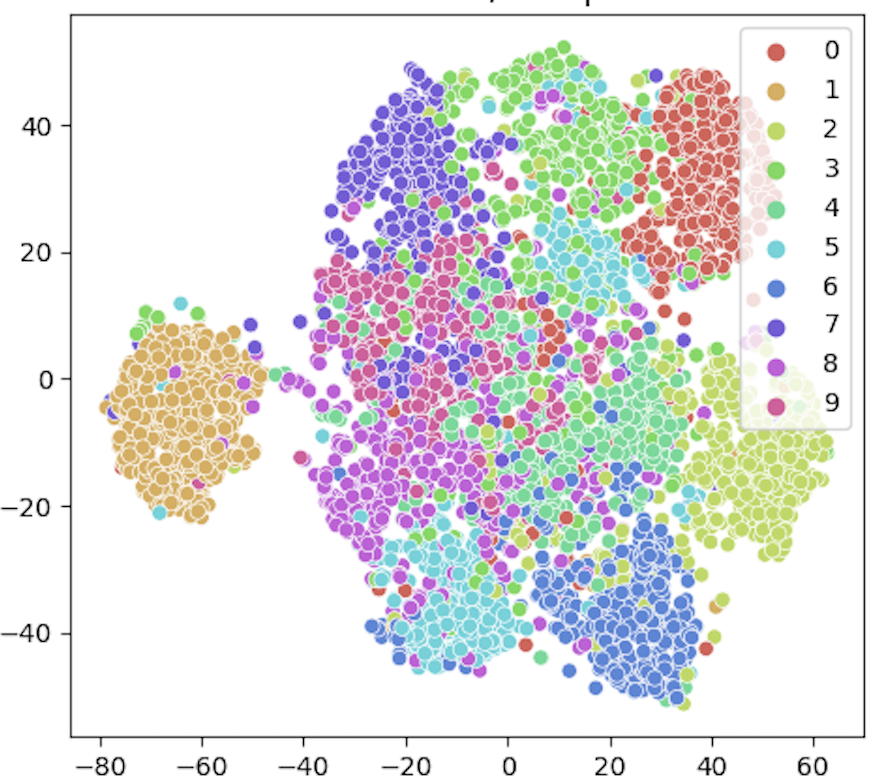}
    \end{subfigure}
    \begin{subfigure}[t]{0.15\textwidth}
        \centering
        \includegraphics[height=0.9in]{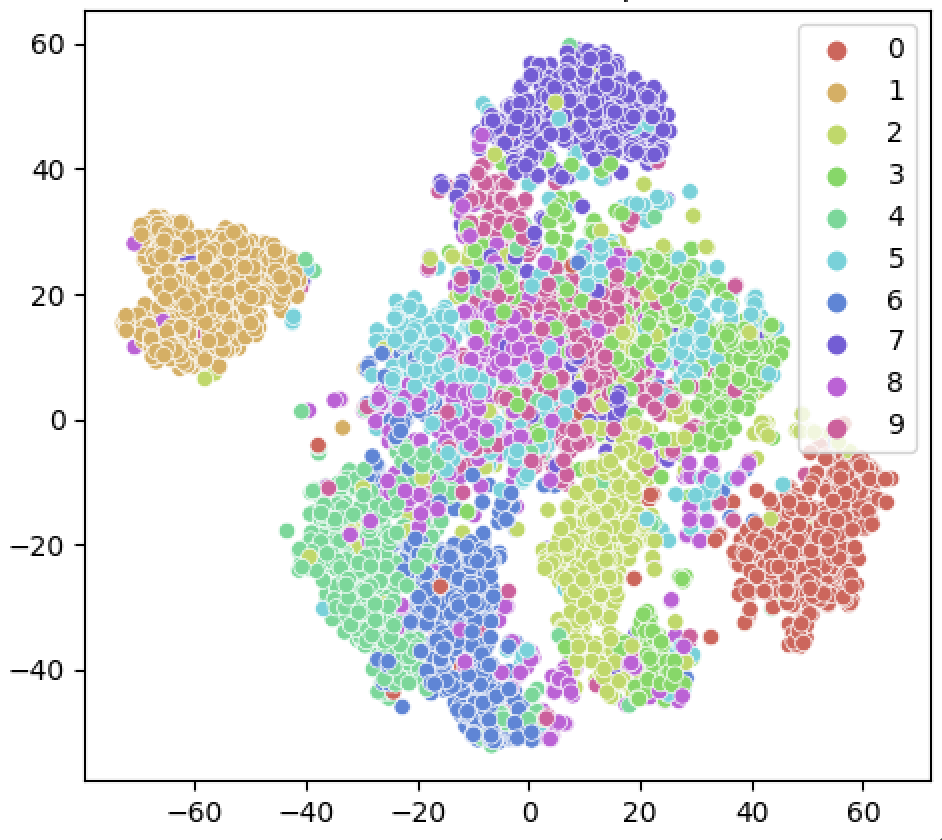}
    \end{subfigure} \\
    \vspace{-2mm}
    \caption{\small t-SNE plots on the evaluation dataset at the start (left), middle (middle) and end (right) of training on sequential data streams on MNIST.}
    \vspace{-2mm}
    \label{fig:tsne-seq}
\end{figure}

To clearly visualize the challenges of learning from sequential incremental input versus \textit{iid} input, we depict the t-SNE plots on the feature space using the evaluation dataset during training \Method. The colors indicate ground-truth class labels. As shown in Figure~\ref{fig:tsne-iid}, under \textit{iid} data streams on MNIST, all classes are quickly separated as the middle-stage t-SNE plot already demonstrates the distinguished class distribution in the feature space. On the contrary, due to the lack of labels and balanced data input, distinguishing and memorizing various classes under class-incremental input is much more difficult as shown in Figure~\ref{fig:tsne-seq}. \Method is able to extract obvious class patterns and discriminate one class versus the others by the $k$NN classifier.

\section{Time Complexity of \Method} 
\label{sec:compute-complexity}

\textbf{Time complexity of loss functions.} We analyze the computation complexity of \Method and compare with state-of-the-art lifelong learning method.

\begin{itemize}[itemsep=-1.5mm]
    \item \textbf{Co2L}~\cite{cha2021co2l} is the state-of-the-art supervised lifelong learning method using contrastive loss and forgetting loss. Both losses depend on the pairwise similarity between all streaming and memory representations. Hence after the forward propagation, the computation complexity of computing the losses is $O((m+n)^2)$, where $m$ and $n$ refer to the memory and streaming batch size respectively.
    \item \textbf{\Method} utilizes the pseudo-contrastive loss and forgetting loss, both based on pairwise similarity and the computation can be reused. Therefore, the computation complexity to compute the losses in \Method is the same as Co2L, both being $O((m+n)^2)$. Moreover, \Method consumes less time and resource than CaSSLe without the predictor.
\end{itemize}

We measure the execution time per batch on a Linux desktop with Intel Core i7-8700 CPU
at 3.2 GHz and 16 GB RAM, and a NVIDIA GeForce 3080Ti GPU. The settings are the same as the implementation details in Section~\ref{sec:implementation}.
The results in Table~\ref{tbl:comp-time} show that \Method consumes nearly the same time as Co2L. The computation time of Co2L and \Method is directly affected by the combined batch size $m+n$, which supports our analyses. 

\begin{table}[h]
\small
\centering
\caption{\small Average computation time (in seconds) of losses per batch in Co2L, CaSSLe and \Method on CIFAR-10, using various batch sizes.}
\vspace{-2mm}
\label{tbl:comp-time}
\begin{tabular}{c|c|c|c} 
\toprule
\small
$n$ & $m$ & \multicolumn{2}{c}{Time (s)} \\ \cline{2-4}
& & Co2L & \Method \\ \hline
128 & 128 & 0.050 & 0.051 \\
64 & 128 & 0.034 & 0.035 \\
128 & 64 & 0.034 & 0.035 \\ 
\bottomrule
\end{tabular}
\vspace{-2mm}
\end{table}

\textbf{Time complexity of memory update.} \Method employs the PSA to select a uniformly distributed subset. We measure the execution time per memory update of random selection, KMeans-based selection, MinRed~\cite{purushwalkam2022challenges} and PSA on the same machine. The results are summarized in Table~\ref{tbl:mem-time}. The configurations are the same as described in Section~\ref{sec:implementation}.
PSA is only slower than the random baseline and executes faster than KMeans-based selection and MinRed. The KMeans-based selection performs KMeans clustering on all latent features and then runs a random update within each cluster. 
We implement KMeans using the \texttt{scikit-learn} library~\cite{scikit-learn} with $k$ equal to the ground-truth number of classes. 
In our setting, KMeans is not ideal as it not only uses prior knowledge of the number of class, but is not computationally efficient due to its iterative nature.
MinRed as a greedy heuristic needs to evaluate all candidates in a sample-by-sample manner. In our implementation, MinRed is 10\% slower than PSA.

While the memory update seems to take much longer time compared to computing loss values, we remind the reader that all above memory selection mechanisms are deployed on CPU, and do not utilize the acceleration capability of GPU. In the future, we plan to re-implement the code to convert to a GPU version.

\begin{table}[h]
\small
\centering
\caption{\small Average computation time (in seconds) per memory update on CIFAR-10, using various memory update policies.}
\vspace{-2mm}
\label{tbl:mem-time}
\begin{tabular}{c|c|c|c|c} 
\toprule
\small
& random & KMeans & MinRed~\cite{purushwalkam2022challenges} & PSA \\ \hline
Time (s) & 0.40 & 1.51 & 1.05 & 0.95 \\
\bottomrule
\end{tabular}
\vspace{-2mm}
\end{table}


\newpage

